\documentclass[final,authoryear,5p,times,twocolumn]{elsarticle}

\usepackage{graphicx}
\usepackage{subfig}
\usepackage[cmex10]{amsmath}
\usepackage{amssymb}
\usepackage{multirow}
\usepackage{url}
\usepackage{placeins}
\usepackage{floatrow}
\usepackage{color}
\usepackage{flushend}

\usepackage{tabularx}
\usepackage{array}
\usepackage{rotating}
\definecolor{darkgreen}{rgb}{0,.5,0}
\definecolor{purple}{rgb}{.5,0,.5}

\newcommand{\changed}[1]{\textcolor{black}{#1}}
\newcommand{\added}[1]{\textcolor{black}{#1}}
\newcommand{\shrunk}[1]{\textcolor{black}{#1}}

\newcommand{\argmin}{\mathop{\mathrm{argmin}}}

\newcolumntype{Y}{>{\centering\arraybackslash}X}
\newcommand\RotText[1]{\rotatebox{90}{\parbox{2.2cm}{\centering#1}}}

\journal{Medical Image Analysis}

\begin{document}

\begin{frontmatter}



\title{Monitoring Tool Usage in \changed{Surgery} Videos using Boosted Convolutional and Recurrent Neural Networks}

\author[label1]{Hassan~Al~Hajj}
\author[label2,label1]{Mathieu~Lamard}
\author[label3,label1]{Pierre-Henri~Conze}
\author[label2,label1,label4]{B\'eatrice~Cochener}
\author[label1]{Gwenol\'e~Quellec\corref{cor1}}
\ead{gwenole.quellec@inserm.fr}
\cortext[cor1]{LaTIM - IBRBS - CHRU Morvan - 12, Av. Foch\\29609 Brest CEDEX - FRANCE\\Tel.: +33 2 98 01 81 29 / Fax: +33 2 98 01 81 24}
\address[label1]{Inserm, UMR 1101, Brest, F-29200 France}
\address[label2]{Univ Bretagne Occidentale, Brest, F-29200 France}
\address[label3]{Institut Mines-T\'el\'ecom Atlantique, Brest, F-29200 France}
\address[label4]{Service d'Ophtalmologie, CHRU Brest, Brest, F-29200 France}

\begin{abstract}
This paper investigates the automatic monitoring of tool usage during a surgery, with potential applications in report generation, surgical training and real-time decision support. \changed{Two surgeries are considered: cataract surgery, the most common surgical procedure, and cholecystectomy, one of the most common digestive surgeries. Tool usage is monitored in videos recorded either through a microscope (cataract surgery) or an endoscope (cholecystectomy).} Following state-of-the-art video analysis solutions, each frame of the video is analyzed by convolutional neural networks (CNNs) whose outputs are fed to recurrent neural networks (RNNs) in order to take temporal relationships between events into account. Novelty lies in the way those CNNs and RNNs are trained. Computational complexity prevents the end-to-end training of ``CNN+RNN'' systems. Therefore, CNNs are usually trained first, independently from the RNNs. This approach is clearly suboptimal for surgical tool analysis: many tools are very similar to one another, but they can generally be differentiated based on past events. CNNs should be trained to extract the most useful visual features in combination with the temporal context. A novel boosting strategy is proposed to achieve this goal: the CNN and RNN parts of the system are simultaneously enriched by progressively adding weak classifiers (either CNNs or RNNs) trained to improve the overall classification accuracy. \changed{Experiments were performed in a dataset of 50 cataract surgery videos, where the usage of 21 surgical tools was manually annotated, and a dataset of 80 cholecystectomy videos, where the usage of 7 tools was manually annotated. Very good classification performance are achieved in both datasets: tool usage could be labeled with an average area under the ROC curve of $A_z = 0.9961$ and $A_z = 0.9939$, respectively, in offline mode (using past, present and future information), and $A_z = 0.9957$ and $A_z = 0.9936$, respectively, in online mode (using past and present information only).}
\end{abstract}
\begin{keyword}
  \changed{cataract and cholecystectomy surgeries} \sep tool usage monitoring \sep video analysis \sep \changed{Convolutional and Recurrent} Neural Networks \sep boosting
\end{keyword}

\end{frontmatter}

\section{Introduction}
\label{sec:Introduction}

With the emergence of imaging devices in the operating room, the automated analysis of videos recorded during the surgery is becoming a hot research topic. \changed{In particular, videos can be used to monitor the surgery, for instance by recognizing} which surgical tools are being used at every moment. \changed{An immediate application of surgery monitoring is report generation. If automatic reports are available for many surgeries, then the automatic analysis of these reports can help optimize the surgical workflow or evaluate surgical skills. Additionally, if we are able to generate such a report in real-time, during a surgery, then we could compare it with previous reports to generate warnings, if we recognize patterns often leading to complications, or recommendations, to help younger surgeons emulate more experienced colleagues based on their surgical reports \citep{quellec_real-time_2015}.} \added{With adequate image analysis techniques, tool usage could be monitored reliably in tool-interaction videos, such as endoscopic videos (in laparoscopic or retinal surgeries) or microscopic videos (in anterior eye segment surgeries). In the simplest scenario, we can consider that a tool is being used if it is visible in these videos. In a more advanced scenario, we can consider that it is in use if it is in contact with the tissue (as opposed to approaching the tissue, waiting to be used, etc.).} Therefore, several tool detection techniques \added{for tool-interaction videos} have been proposed in recent years \citep{bouget_vision-based_2017}. To compare these techniques, \changed{two tool detection challenges were organized recently. A first challenge,} organized at the M2CAI 2016 workshop,\footnote{\url{http://camma.u-strasbg.fr/m2cai2016/index.php/tool-presence-detection-challenge-results/}} \changed{relied on endoscopic videos of cholecystectomy operations performed laparoscopically.} \added{We organized a second challenge for cataract surgery, the most common surgical procedure worldwide \citep{trikha_journey_2013}.\footnote{\url{https://cataracts.grand-challenge.org/}} It relied on videos recorded through a surgical microscope.} Following the trend in medical image and video analysis \citep{shen_deep_2017}, the best solutions \added{of both challenges} all relied on convolutional neural networks (CNNs) \citep{raju_m2cai_2016, sahu_tool_2016, twinanda_endonet:_2017, zia_fine-tuning_2016, roychowdhury_identification_2017, hu_surgical_2017, marsalkaite_towards_2017}.

Compared to other computer vision tasks, surgical tool usage annotation has several specificities. First, as opposed to many computer vision tasks, including the popular ImageNet visual recognition challenges,\footnote{\url{http://www.image-net.org/challenges/LSVRC/2017/index.php}} the problem at hand is not multiclass classification (one correct label per image among multiple classes), but rather multilabel classification (multiple correct labels per image): the number of tools being used in each image varies (from zero to three in cataract surgery \added{for instance}). Therefore, multilabel CNNs should be used. Second, taking the temporal sequencing into account is important: knowing which tools have already been used since the beginning of the surgery greatly helps recognize which tools are currently being used. Therefore, multilabel recurrent neural networks (RNNs) \citep{hochreiter_long_1997} may also be used advantageously. In fact, recent machine learning competitions clearly show that ensembles of CNNs outperform single CNNs \citep{russakovsky_imagenet_2015}: multiple CNNs with different architectures are generally trained independently and their outputs are combined afterward using standard machine learning algorithms (decision trees, random forests, multilayer perceptrons, etc.). However, this simple strategy is suboptimal since difficult samples may be misclassified by all CNNs. And there are many difficult samples to classify in surgery videos: in particular, many tools resemble one other (e.g. two types of cannulae in cataract surgery). Building the ensemble of CNNs using a boosting meta-algorithm \citep{freund_decision-theoretic_1997} can theoretically design CNNs focusing specifically on challenging samples. Boosting an ensemble of RNNs would also make sense as there are difficult samples along the time dimension as well: in particular, some tools or tool usage sequences are very rare and temporal sequencing algorithms tend to misclassify those rare cases. Therefore, we propose to jointly boost an ensemble of CNNs and an ensemble of RNNs for automatic tool usage annotation in surgery videos. In the same way as CNN boosting (or RNN boosting) allows various CNNs (or RNNs) to be complementary, this general boosting solution allows CNNs to be complementary with RNNs. In that sense, it approximates the end-to-end training of a ``CNN+RNN'' network, which is theoretically ideal but not computationally tractable.

The remainder of this paper is organized as follows. Section \ref{sec:StateArt} reviews the state of the art \changed{of video analysis, and surgery video analysis in particular}. \changed{Sections \ref{sec:CNNRNNNetworks} and \ref{sec:BoostedCNNRNNNetworks} describe the proposed solution.} Section \ref{sec:datasets} presents \changed{the video datasets} and section \ref{sec:Experiments} reports the experiments performed on that dataset. We end with a discussion and conclusions in section \ref{sec:DiscussionConclusions}.

\section{State of the Art}
\label{sec:StateArt}

\subsection{Deep Learning for Video Analysis}
\label{sec:DeepLearningVideoAnalysis}

\shrunk{The automatic analysis of dynamic scenes through deep learning has become a very hot research topic \citep{simonyan_two-stream_2014, wang_beyond_2017, donahue_long-term_2017}. Different strategies have been proposed for this task. A first strategy is to regard videos or video portions as 3-D images and therefore analyze them with 3-D CNNs \citep{ji_3d_2013}, which is computationally expansive. A second strategy is to analyze 2-D images as well as the optical flow between consecutive images \citep{simonyan_two-stream_2014}, with the disadvantage of only modeling short-term relationships between images. A third strategy is to combine a CNN, analyzing 2-D images, with a RNN analyzing the temporal sequencing \citep{donahue_long-term_2017}. The main advantage of this ``CNN+RNN'' approach, which is now the leading video analysis solution, is that long-term relationships between events can be taken into account efficiently. One application of ``CNN+RNN'' models, which is particularly relevant for our study, is video labeling: the goal is to assign one class label to each frame inside a video \citep{singh_multi-stream_2016, khorrami_how_2016}. Medical applications of this research, ranging from gait analysis \citep{feng_learning_2016} to surgery monitoring \citep{bodenstedt_unsupervised_2017, twinanda_single-_2016}, are starting to emerge.}

\subsection{\added{Temporal Analysis of Surgery Videos}}

\changed{In the context of surgical workflow analysis, solutions have been proposed to recognize surgical phases in surgery videos} \citep{lalys_surgical_2014, charriere_real-time_2017}. \added{In \citet{primus_frame-based_2018}, phases are recognized using one CNN processing the visual content of one frame plus the relative timestamp of that frame.} However, most solutions rely on statistical models, such as Hidden Markov Models (HMMs) \citep{cadene_m2cai_2016}, Hidden semi-Markov Models \citep{dergachyova_automatic_2016, tran_phase_2017}, Hierarchical HMMs \citep{twinanda_endonet:_2017}, Linear Dynamical Systems \citep{zappella_surgical_2013, tran_phase_2017} or Conditional Random Fields \citep{tao_surgical_2013, quellec_real-time_2014, lea_learning_2016}. Recently, solutions based on RNNs have also been proposed \citep{jin_endorcn:_2016, bodenstedt_unsupervised_2017, twinanda_single-_2016}. Following the state-of-the-art video analysis strategy, these RNNs process instant visual features extracted by a CNN from images. In particular, \citet{jin_endorcn:_2016} applied a \changed{``CNN+RNN''} network to a small sliding window of three images. \citet{bodenstedt_unsupervised_2017} applied a \changed{``CNN+RNN''} network to larger sliding windows and copy the internal state of the network between consecutive window locations. As for \citet{twinanda_single-_2016}, they applied a \changed{``CNN+RNN''} network to full videos. Interestingly, the CNN proposed by \citet{twinanda_single-_2016}, namely EndoNet, detects tools as an intermediate step. A challenge on surgical workflow analysis was also organized at M2CAI 2016:\footnote{\url{http://camma.u-strasbg.fr/m2cai2016/index.php/workflow-challenge-results/}} two of the top three solutions relied on RNNs \citep{jin_endorcn:_2016, twinanda_single-_2016}. It should be noted that successful works on the analysis of kinematics surgery data have also been reported, using a \changed{RNN} \citep{dipietro_recognizing_2016} or a CNN along the temporal dimension \citep{lea_temporal_2016}. In all these works, statistical models or RNNs were used to label surgical activities and phases. Given the strong correlation between surgical activities and tool usage, they can be expected to improve tool recognition as well.

\subsection{\changed{Deep Learning for Surgical Tool Detection}}

As evidenced by the M2CAI 2016 \added{and CATARACTS 2017} challenges, the state-of-the-art algorithms for tool detection in surgery videos are CNNs. The best solutions of \changed{these challenges} rely on a transfer learning strategy: well-known CNNs trained to classify still images in the ImageNet dataset were fine-tuned on images extracted from surgery videos. \changed{For M2CAI 2016}, \citet{sahu_tool_2016} and \citet{twinanda_endonet:_2017} fine-tuned AlexNet \citep{krizhevsky_imagenet_2012}, \citet{raju_m2cai_2016} fine-tuned GoogleNet \citep{szegedy_going_2015} and VGG-16 \citep{simonyan_very_2015}, and \citet{zia_fine-tuning_2016} fine-tuned AlexNet, VGG-16 and Inception-v3 \citep{szegedy_rethinking_2015}. \added{For CATARACTS, \citet{roychowdhury_identification_2017} fine-tuned Inception-v4 \citep{szegedy_inception-v4_2017}, ResNet-50 \citep{he_deep_2016} and two NASNet-A instances \citep{zoph_learning_2017}, \citet{hu_surgical_2017} fine-tuned ResNet-101 and DenseNet-169 \citep{huang_densely_2017}, and \citet{marsalkaite_towards_2017} fine-tuned four ResNet-50 instances}. \added{Training a CNN proved challenging due to highly frequent tool co-occurrences: a solution based on label-set sampling has been proposed by \citet{sahu_addressing_2017} to reduce this bias.} Note that temporal information is not exploited in these solutions, \added{with a few exceptions presented hereafter \citep{sahu_addressing_2017, marsalkaite_towards_2017, al_hajj_surgical_2017, mishra_learning_2017, roychowdhury_identification_2017}. In \citet{sahu_addressing_2017} and \citet{marsalkaite_towards_2017}, a linear filter is used to smooth CNN predictions from consecutive frames.} \changed{In \citet{al_hajj_surgical_2017}, a CNN processes} short sequences of consecutive images, using the optical flow to register and combine local features from consecutive images. \added{In \citet{mishra_learning_2017}, one RNN processes the outputs of a frame-level CNN inside short sequences of consecutive frames.} Note that long-term relationships between images are not exploited neither in \changed{these four solutions}: the goal is to combine slightly different views on a tool, some of which being affected by motion blur or occlusion. \added{In \citet{roychowdhury_identification_2017}, on the other hand, long-term relationships between images are exploited through a Markov Random Field (MRF) modeling long sequences of approximately 20,000 frames. The drawback is that online video analysis is not possible.}

\subsection{Proposed Solution}
\label{sec:ProposedSolution}

\begin{figure}[!t]
  \begin{center}
    \includegraphics[scale=.425]{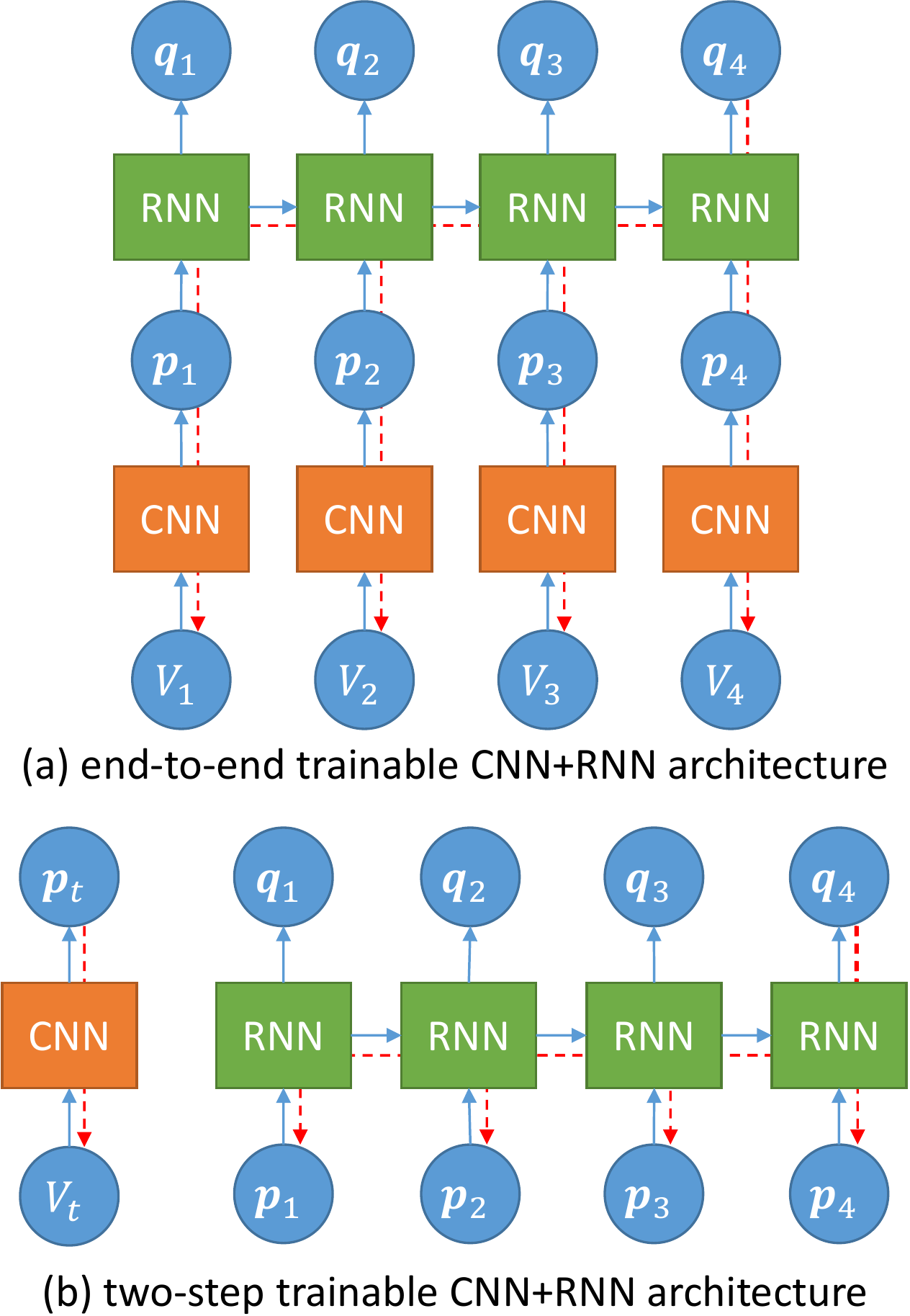}
  \end{center}
  \caption{\shrunk{Training strategies for ``CNN+RNN'' networks. Each green cell represent one RNN cell (or several RNN cells stacked on top of each other in a multi-layer RNN). Each orange cell represents one CNN; $\boldsymbol{p}_t$ and $\boldsymbol{q}_t$ are short notations for $\boldsymbol{p}(V_t)$ and $\boldsymbol{q}(V_t)$, respectively. Two ``CNN+RNN'' training strategies are illustrated in Fig. (a) and (b).} They reveal that the first strategy (a) is not tractable: backpropagating errors at time index $t$ involves $t$ backpropagations through the CNN, as illustrated in red for $t=4$.}
  \label{fig:TrainingStrategiesCNNRNNNetworks}
\end{figure}

In this paper, we propose to design ``CNN+RNN'' networks, the state-of-the-art video analysis framework, for the task of automatic tool usage annotation. Due to the specific challenges of this task, namely the similarity between some tools and the rarity of some tool usages, we propose to apply the boosting principle to both the CNN part and the RNN part of the network, in a novel and unified manner. Besides addressing the previously mentioned difficult cases, the proposed framework has multiple advantages: 1) it can be used to select the \changed{network architectures automatically}, an open problem in deep learning, and 2) it can improve the complementarity of CNNs and RNNs, an unsolved problem in ``CNN+RNN'' models for which end-to-end learning is not tractable (see Fig. \ref{fig:TrainingStrategiesCNNRNNNetworks}). Section \ref{sec:CNNRNNNetworks} briefly describes the networks considered in this paper and the related challenges. Section \ref{sec:BoostedCNNRNNNetworks} describes the boosting algorithm proposed to address those challenges. \shrunk{The proposed solution has several novelties. First, the use of CNN boosting and RNN boosting for medical images or videos is novel. Second, the \changed{data-driven design of a} CNN or CNN ensemble to be used as input for an RNN or RNN ensemble (through boosting --- see section \ref{sec:BoostingCNNsInsideCNNRNNNetwork}) has never been studied before.}

\section{``CNN+RNN'' Networks}
\label{sec:CNNRNNNetworks}

\subsection{Notations}
\label{sec:Notations}

Let $\Theta$ denote a set of surgical tools \changed{whose usage should be monitored} in videos. Let $\mathcal{D}$ denote a collection of training videos \changed{and let} $V_t$ denote the $t$-th frame in video $V\in\mathcal{D}$. Let $\delta(V_t, \theta)\in \left\{ -1, 1 \right\}$ denote the binary label assigned to frame $V_{t}$ for tool $\theta\in\Theta$: this label indicates whether or not tool $\theta$ is being used in frame $V_t$. \changed{We are addressing a multilabel classification problem, so $0 \leq \sum_\theta{\delta(V_t, \theta)} \leq |\Theta|$. In contrast, $\sum_\theta{\delta(V_t, \theta)} = 1$ in a multiclass classification problem.}

Neural networks considered in this paper consist of one or several CNNs working in parallel: this set of CNNs is referred to as the ``CNN block''. Let $\boldsymbol{p}(V_t) = \left\{ p(V_t, \theta) \in [0; 1], \theta\in\Theta \right\}$ denote the instant predictions computed by the CNN block for frame $V_t$. Some of the neural networks considered in this paper also contain one or several RNNs working in parallel: this set of RNNs is referred to as the ``RNN block''.  Let $\boldsymbol{q}(V_t) = \left\{ q(V_t, \theta) \in [0; 1], \theta\in\Theta \right\}$ denote the context-aware predictions computed by the RNN block for frame $V_t$.

\subsection{RNNs \added{Processing CNN Predictions}}

\shrunk{A recurrent neural network (RNN) is a neural network that takes a sequence of observations at the input and produces a sequence of predictions at the output \citep{hochreiter_long_1997}. In this paper, the input sequence is $\left\{ \boldsymbol{p}(V_t) | t = 1 .. |V| \right\}$, i.e. the predictions of the CNN block for each frame in a video. The output sequence is $\left\{ \boldsymbol{q}(V_t) | t = 1 .. |V| \right\}$. The network is structured in such a way that the prediction vector $\boldsymbol{q}(V_t)$ depends on feature vector $\boldsymbol{p}(V_t)$, but also on all previous feature vectors $\boldsymbol{q}(V_u)$, $u < t$. This behavior is achieved by 1) connecting each input element $\boldsymbol{p}(V_t)$ to a group of neurons $C_t$ called ``cell'', 2) connecting $C_t$ to the output element $\boldsymbol{q}(V_t)$ and 3) connecting $C_t$ to the next cell $C_{t+1}$. Weights are shared across all cells. The most popular cells are Long Short-Term Memory (LSTM) cells \citep{hochreiter_long_1997}: they include a ``forgetting'' mechanism preventing backpropagated errors from vanishing or exploding in long sequences. More recently, Gated Recurrent Units (GRU) were proposed by \citet{cho_properties_2014}: the labeling performance of these lower-complexity cells is often comparable with LSTM.}

\begin{figure}[!t]
  \begin{center}
    \includegraphics[scale=.425]{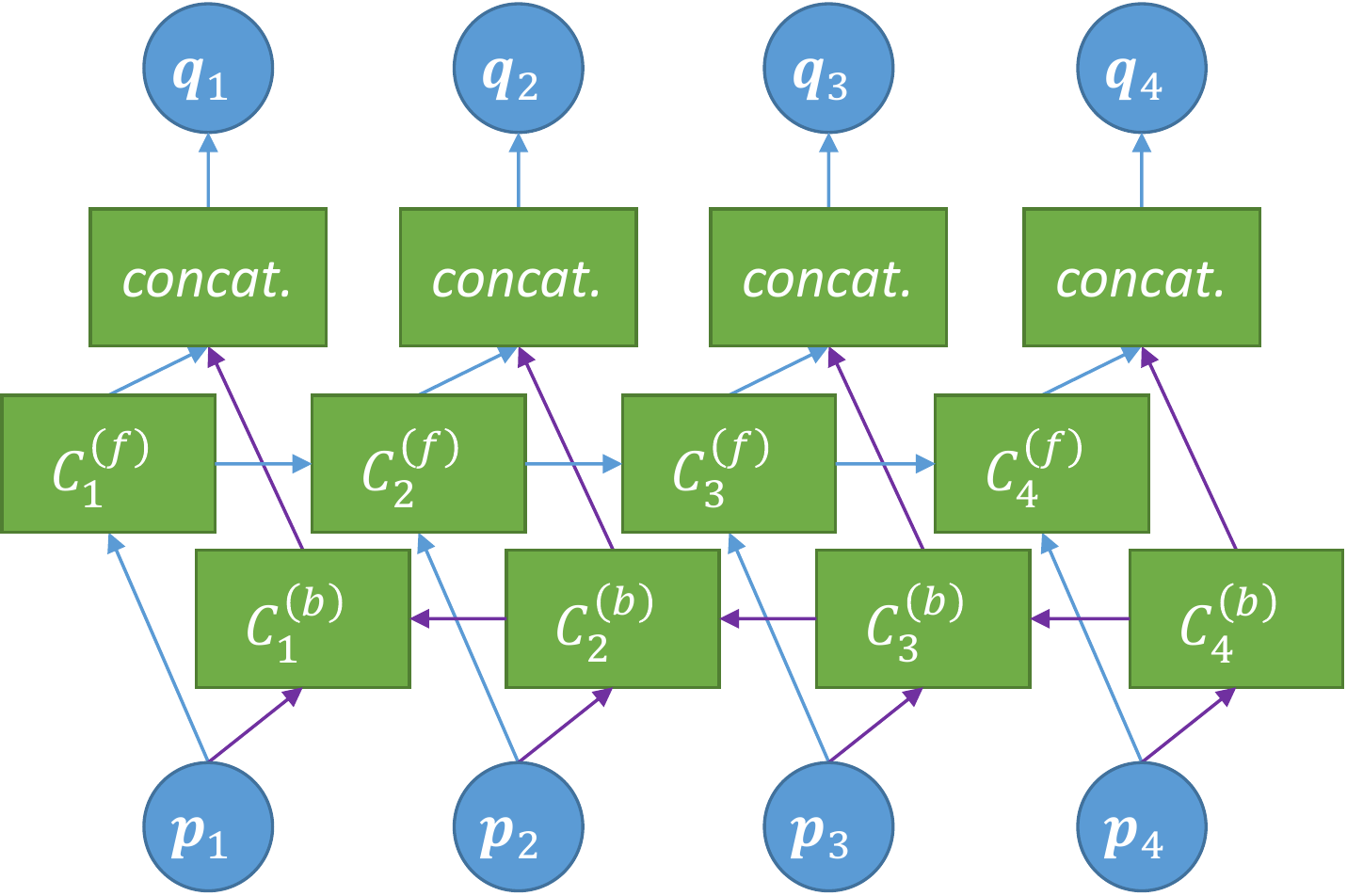}
  \end{center}
  \caption{\shrunk{Bidirectional RNN networks. Three elements are defined at each timestamp: 1) a forward RNN cell (or stack of RNN cells), 2) a backward RNN cell (or stack of RNN cells) and 3) a fusion part, which concatenates their outputs. The purple arrows represent information propagated backward in time.}}
  \label{fig:BidirectionalRNNNetworks}
\end{figure}

\shrunk{A multi-layer extension was proposed for RNNs. In this extension, each timestamp $t$ is associated with multiple cells $C_{i, t}$, where $i=1..n$ is the layer index. At each timestamp $t$, $\boldsymbol{p}(V_t)$ is connected to $C_{1, t}$, $C_{i, t}$ is connected to $C_{i + 1, t}$ for $i=1..n-1$, and $C_{n, t}$ is connected to $\boldsymbol{p}(V_t)$. In each layer $i$, $C_{i, t}$ is connected to $C_{i, t+1}$. Weights are shared across all cells in the same layer. A bidirectional extension was also proposed for RNNs \citep{schuster_bidirectional_1997}. In this extension, illustrated in Fig. \ref{fig:BidirectionalRNNNetworks}, two independent RNNs are defined: in one of them, information flows from timestamp $t$ to timestamp $t+1$; in the other one, information flows from timestamp $t$ to timestamp $t-1$. Their outputs are concatenated and connected to the output sequence. \added{The performance of bidirectional RNNs, which take advantage of past and future information, is generally higher. The drawback is of course that online video labeling is not possible.}}

\subsection{RNNs on Long Video Sequences}
\label{sec:RNNsLongVideoSequences}

In the literature, RNNs are generally trained using video sequences consisting of a few dozen frames \changed{at most \citep{chen_action_2017, gammulle_two_2017, mishra_learning_2017}}. In contrast, analyzing all frames of full surgery videos requires the analysis of much longer sequences: for instance, there are at least 10,000 frames per video sequence in our cataract surgery videos (see section \ref{sec:CATARACTSDataset}). Training long-term relationships with RNNs is more computationally intensive using long sequences, so we propose to analyze shorter sequences. In that purpose, $M$ subsampled versions of each original sequence $V$, denoted by $V^{(m)}$, $m=1..M$, are generated as follows:
\begin{equation}
  V^{(m)} = \left\lbrace V_u \; | \; u = m + t M, t \in \mathbb{N}^*, u \leq |V| \right\rbrace \; .
\end{equation}
During training, this results in a novel kind of data augmentation \citep{shen_deep_2017}: the number of training sequences increases artificially. For simplicity, $\left\lbrace V^{(m)} \; | \; V \in \mathcal{D}, m = 1..M \right\rbrace$ is denoted by $\mathcal{D}$ in the remainder of this paper. During testing, each of the $M$ subsequences of $V$ are analyzed independently and the final prediction sequence for $V$ is obtained by interleaving the resulting $M$ prediction sequences. \added{The resulting prediction sequence is further processed by median filters to blend subsequences: a filter of radius $R_\theta$ is used for each tool-specific channel of the sequence.}

\subsection{\changed{Training Complexity for} ``CNN+RNN'' Networks}

Because CNNs and RNNs are integrated into the same network, it would make sense to train the entire network from end to end, so that features extracted by the CNNs are as relevant as possible to the RNNs that process them further. However, as illustrated in Fig. \ref{fig:TrainingStrategiesCNNRNNNetworks}, the complexity of the learning process is very high. The error measured for each prediction $q(V_t, \theta)$ is backpropagated to $\boldsymbol{p}(V_t)$ but also to all $\boldsymbol{p}(V_u)$ (such as $u \leq t$, in unidirectional networks). Errors computed for each $\boldsymbol{p}(V_u)$ are backpropagated further towards $V_u$.

The vast majority of weights in a ``CNN+RNN'' network are in the CNNs. Therefore, the cost of backpropagating an error measured for one timestamp $t$ to all frames $V_u$ in the video sequence (such as $u \leq t$, in unidirectional networks) is very high. As a consequence, a two-step training process is always preferred in the literature (see section \ref{sec:DeepLearningVideoAnalysis}). A CNN is trained first: errors measured for one timestamp $t$ are only backpropagated to $V_t$. Then, a RNN is trained: errors measured for one timestamp $t$ are backpropagated to all $\boldsymbol{p}(V_u)$ (such as $u \leq t$, in unidirectional networks) without affecting the CNN weights. Given the number of weights in a RNN, this process is tractable. We propose a solution based on boosting that is able to improve the CNN block after or while training the RNN block, in order to achieve the desirable properties of end-to-end training, but at a reasonable computational cost.

\begin{figure}[!t]
  \begin{center}
    \includegraphics[scale=.425]{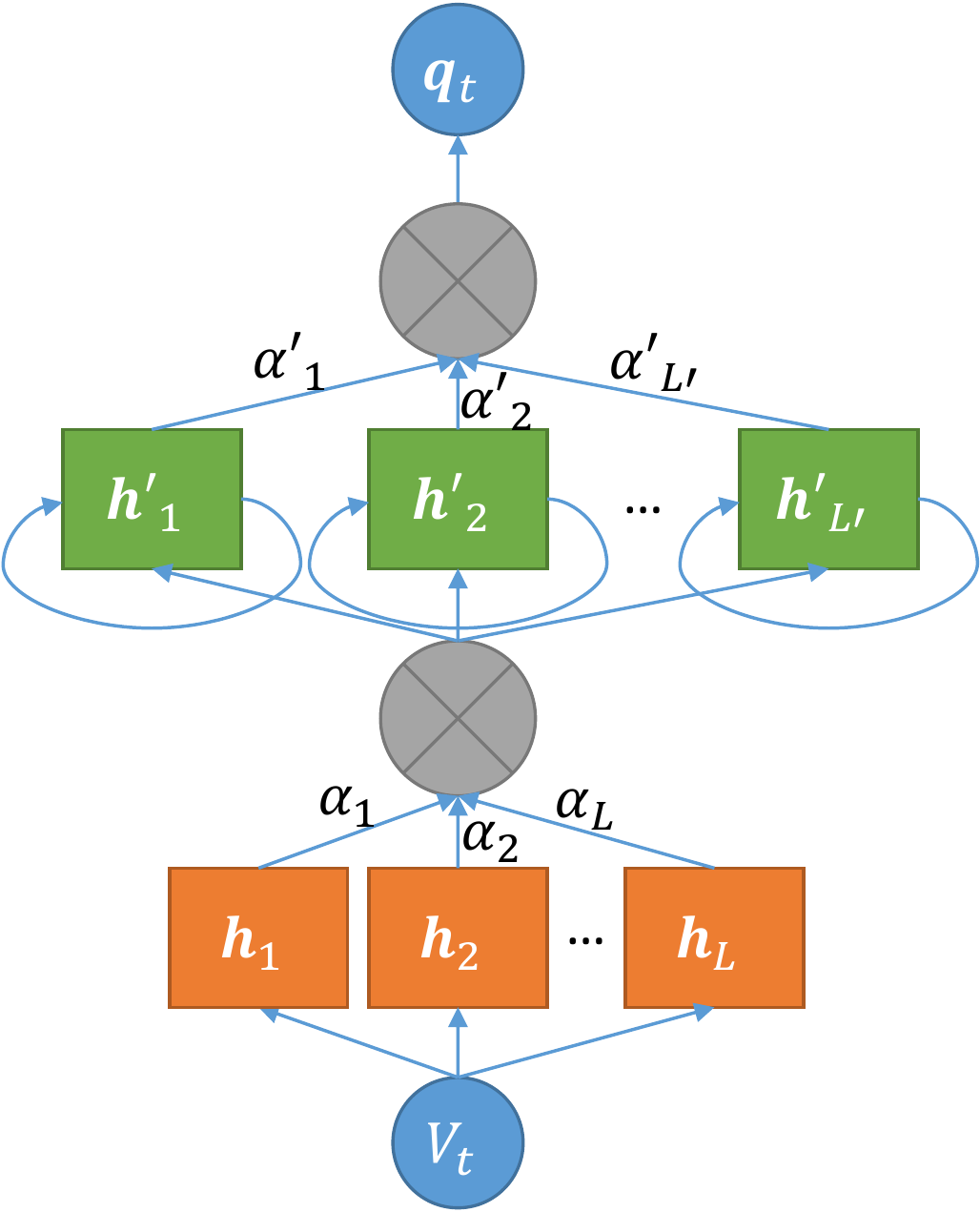}
  \end{center}
  \caption{\shrunk{Boosted ``CNN+RNN'' network (unidirectional version). The $\otimes$ symbol represents the sigmoid operator applied to the weighted sum of the inputs.}}
  \label{fig:BoostedCNNRNNNetworks}
\end{figure}

\section{Boosted ``CNN+RNN'' Networks (see Fig. \ref{fig:BoostedCNNRNNNetworks})}
\label{sec:BoostedCNNRNNNetworks}

\subsection{\added{Context}}

\shrunk{Recent boosting algorithms, such as AnyBoost \citep{mason_boosting_1999} and \citet{friedman_greedy_2001}'s Gradient Boosting Machines (GBM), are formulated as a gradient descent optimization, which integrates nicely with the way neural networks are trained. When CNNs or RNNs are used as weak learners, the boosting meta-algorithm controls the loss function used to train these learners. Typically, training samples with large classification errors are assigned a larger weight in the updated loss function. A few authors thus used CNNs as weak learners for AnyBoost \citep{moghimi_boosted_2016} or GBM \citep{zhang_scene_2016, walach_learning_2016}.} A boosting algorithm based on GBM \citep{friedman_greedy_2001} is proposed in this section to design either a CNN block or an RNN block. The same algorithm is used for CNN boosting in \changed{RNN-free} networks and for RNN boosting in ``CNN+RNN'' networks. To ensure the complementarity of the CNN and RNN blocks in ``CNN+RNN'' networks, an improved criterion is proposed for CNN boosting in such networks (see section \ref{sec:BoostingCNNsInsideCNNRNNNetwork}). How to design an adequate neural network architecture for a given classification problem remains an open question. So, generalizing \citet{gao_convolutional_2016}, multiple architectures of neural networks (CNNs or RNNs) are considered in this study; let $\mathcal{H}$ denote the set of (CNN or RNN) architectures.

\subsection{Gradient Boosting Machine}
\label{sec:MultilabelGradientBoostingMachine}

\shrunk{The purpose of GBM is to build a strong learner $\boldsymbol{H}_L$ by linearly combining multiple weak learners $\boldsymbol{h}_l \in \mathcal{H}$, l=1..L, with weights $\alpha_l$. Let $\boldsymbol{h}_l(x) = \left\{ h_l(x, \theta), \theta\in\Theta \right\}$ denote the predictions of $\boldsymbol{h}_l$ for some input $x$. The predictions of the strong learner for $x$ are given by:
\begin{equation}
\label{eq:logitSum}
  \boldsymbol{H}_L(x) = \sum_{l=1}^L{\alpha_l \boldsymbol{h}_l(x)} \;.
\end{equation}
These predictions are mapped to probabilities using the sigmoid function $\sigma$: $p_L(x, \theta) = \sigma(H_L(x, \theta))$ in CNN boosting, $q_L(x, \theta) = \sigma(H_L(x, \theta))$ in RNN boosting.
Weak learners are added sequentially in order to minimize the negative log-likelihood \citep{friedman_greedy_2001}:
\begin{equation}
  \label{eq:nll}
  \begin{array}{rl}
   \displaystyle \mathcal{L}(\boldsymbol{h}) = -\sum_{\theta\in \Theta} & \displaystyle \left[ \sum_{x, \delta(x, \theta)=1}{\log \sigma(h(x, \theta))} \right.\\
   \displaystyle                                                                                        & + \displaystyle  \left. \sum_{x, \delta(x, \theta)=-1}{\log\left[1 - \sigma(h(x, \theta)) \right]} \right] \;,
  \end{array}
\end{equation}
where $\delta(x, \theta)$ is the binary label assigned to $x$ for tool $\theta$ (see section \ref{sec:Notations}). At each boosting iteration $L+1$, all weak learners $\boldsymbol{h} \in \mathcal{H}$ are trained as detailed in sections \ref{sec:lossFunctions} to \ref{sec:trainingNeuralNetworksWeakLearners}. Then, the weak learner $\boldsymbol{h}$ minimizing $\mathcal{L}(\boldsymbol{H}_L+\alpha \boldsymbol{h}), \alpha \geq 0$, is added to the strong classifier:
\begin{equation}
  \label{eq:select}
  \left(\boldsymbol{h}_{L+1}, \alpha_{L+1}\right) = \argmin_{\left(\boldsymbol{h} \in \mathcal{H}, \alpha \geq 0\right)}{\mathcal{L}(\boldsymbol{H}_L+\alpha \boldsymbol{h})} \;.
\end{equation}
Boosting stops when $\mathcal{L}$ stops decreasing.}

\subsection{Loss Function for Boosting Neural Networks}
\label{sec:lossFunctions}

\shrunk{As noted by \citet{friedman_greedy_2001}, the weak learner $h_{L+1}$ selected at boosting iteration $L+1 > 1$ should ideally return values $h_{L+1}(x, \theta)$ proportional to $-\frac{\partial \mathcal{L}(\boldsymbol{H}_L)}{\partial H_L(x, \theta)}$:
\begin{eqnarray}
  \label{eq:ideal}
  h_{L+1}(x, \theta)      &=& \kappa \; \omega_{L+1}(x, \theta), \; \forall x, \; \forall \theta, \; \kappa \in \mathbb{R} \;, \\
  \omega_{L+1}(x, \theta) &=& -\frac{\partial \mathcal{L}(\boldsymbol{H}_L)}{\partial H_L(x, \theta)} \;,
\end{eqnarray}
where the $\omega_{L+1}(x, \theta)$ coefficients, called sample weights, are given by:
\begin{equation}
  \label{eq:valueOmega}
  \omega_{L+1}(x, \theta) = \left\lbrace \begin{array}{ccl}
    1 - \sigma\left(H_L(x, \theta) \right) & \textnormal{if} & \delta(x, \theta) = 1 \\
    - \sigma\left(H_L(x, \theta) \right) & \textnormal{if} & \delta(x, \theta) = -1 \\
  \end{array} \right. \;.
\end{equation}
With that property, the strong learner's loss function would decrease directly towards zero. Neural networks can be trained to solve Eq. (\ref{eq:ideal}) in the least square sense, using $\kappa = 1$ without loss of generality. Therefore, the following quadratic loss function can be used for $L > 0$ \citep{moghimi_boosted_2016}:
\begin{equation}
  \label{eq:quadraticLoss}
  \mathcal{L}_2(\boldsymbol{h}, \boldsymbol{\omega}) = \sum_\theta{\sum_x{  \left( h(x, \theta)-\omega(x, \theta) \right)^2  }} \;.
\end{equation}}

\subsection{Efficiently Training Neural Networks as Weak Learners}
\label{sec:trainingNeuralNetworksWeakLearners}

\shrunk{The proposed solution for training weak learners can be summarized as follows. At iteration 1 ($L = 0$), each weak learner $\boldsymbol{h} \in \mathcal{H}$ is trained to minimize $\mathcal{L}(\boldsymbol{h})$, the negative log likelihood [see Eq. (\ref{eq:nll})]. CNN weights are fine-tuned from a model trained on ImageNet; RNN weights are initialized at random. At iterations $L + 1$, $L > 0$, each weak learner $\boldsymbol{h} \in \mathcal{H}$ is trained to minimize $\mathcal{L}_2(\boldsymbol{h}, \boldsymbol{\omega}_{L+1})$, the quadratic loss function [see Eq. (\ref{eq:quadraticLoss})]. Following \citet{moghimi_boosted_2016}, the neuron weights of $\boldsymbol{h}$ are fine-tuned from neuron weights obtained at the previous boosting iteration. This strategy saves time and also improves performance. Indeed, more and more samples receive marginal weights at each boosting iteration, as the classification error decreases [see Eq. (\ref{eq:valueOmega})]. Therefore, the training set somehow becomes smaller and smaller. The proposed strategy can be regarded as transfer learning from a larger dataset, which is known to be beneficial.}

\subsection{Boosting CNNs inside a ``CNN+RNN'' Network}
\label{sec:BoostingCNNsInsideCNNRNNNetwork}

The boosting solution described in previous sections is suboptimal for CNN boosting in a ``CNN+RNN'' network. Let us assume that one image in a video sequence is wrongly classified by the firstly selected CNN $\boldsymbol{h}_1$. Based on the temporal context, the RNN block \changed{might be able to} correct this classification error. Therefore, building a second CNN $\boldsymbol{h}_2$ for correcting \changed{that error} specifically \changed{might be useless}. Instead, CNNs should be trained to maximize the performance of the ``CNN+RNN'' network as a whole.

Throughout the rest of this paper, let $\boldsymbol{H}'$, $\boldsymbol{h}'$, $\boldsymbol{\alpha}'$ and $L'$ denote respectively the strong learner, the weak learners, their weights and their number in the RNN block, in order to avoid confusion with their counterparts in the CNN block. To achieve the desired behavior, the sample weights $\boldsymbol{\omega}_{L+1}$ should be defined based on $\boldsymbol{q}_{L'}$, the outputs of the RNN block, rather than $\boldsymbol{p}_L$, the outputs of the CNN block: the goal should be to minimize $\mathcal{L}(\boldsymbol{H}_L, \boldsymbol{H'}_{L'})$. In this scenario, $\omega_{L+1}(V_t, \theta)$, the weight assigned to frame $V_t$ and label $\theta \in \Theta$, does not depend solely on instant quantities, namely $\boldsymbol{H}_L(V_t)$ and $\delta(V_t, \theta)$. In bidirectional networks (for offline processing), it depends on all $\left( \boldsymbol{H}_L(V_u), \delta(V_u, \phi) \right)$ pairs, $\phi \in \Theta$. In unidirectional networks, it depends on all pairs such that $u \geq t$. For $L > 0$, \changed{sample} weights become:
\begin{equation}
  \label{eq:valueOmegaPrime}
  \left\lbrace \begin{array}{rcl} \displaystyle \omega_{L+1}(V_t, \theta)
                                                                                       & =        & p_L(V_t, \theta) (1 - p_L(V_t, \theta)) \\
                                                                                       & \times & \displaystyle  \sum_{\phi \in \Theta}\sum_{V_u}{ \Delta^{\delta(V_u, \phi)}(V_t, \theta, V_u, \phi) } \\
                                 \Delta^+(V_t, \theta, V_u, \phi)  & =        & \displaystyle (1 - q_{L'}(V_u, \phi)) \sum_{l = 1}^{L'}{ \alpha'_l \frac{\partial h'_l(V_u, \phi)}{\partial p_L(V_t, \theta)} } \\
                                 \Delta^-(V_t, \theta, V_u, \phi)   & =        & \displaystyle -q_{L'}(V_u, \phi) \sum_{l = 1}^{L'}{ \alpha'_l \frac{\partial h'_l(V_u, \phi)}{\partial p_L(V_t, \theta)} } \\
  \end{array} \right. \;.
\end{equation}
If a unidirectional RNN network is used, then the $\partial h'_l(V_u, \phi) / \partial p_L(V_t, \theta)$ partial derivatives equal zero for all $u < t$. In all other cases, they can be computed automatically by the backpropagation algorithm. Note that the backpropagation algorithm does not compute each $\frac{\partial O_i}{\partial I_j}$ term individually, where $I$ denotes an input tensor whose influence on the output tensor $O$ should be computed. Instead, it computes:
\begin{equation}
  \sum_i{\frac{\partial O_i}{\partial I_j} \nabla_i},
\end{equation}
given a tensor $\nabla$ weighting each coefficient of the output tensor. However, Eq. (\ref{eq:valueOmegaPrime}) can be computed setting:
\begin{itemize}
  \item $O_i = h'_l(V_u, \phi)$, $i = (u, \phi)$,
  \item $I_j = p_L(V_t, \theta)$, $j = (t, \theta)$,
  \item $\nabla_i = 1 - q_{L'}(V_u, \phi)$ or $\nabla_i = q_{L'}(V_u, \phi)$ depending on $\Delta^{\delta(V_u, \phi)}$.
\end{itemize}

\paragraph{Proof for Eq. (\ref{eq:valueOmegaPrime})} In this scenario, the partial derivative of the negative log-likelihood function [see Eq. (\ref {eq:nll})], with respect to $H_L(V_t, \theta)$, is given by:
\begin{equation}
  \begin{array}{rl} \displaystyle \frac{\partial \mathcal{L}(\boldsymbol{H}_L, \boldsymbol{H'}_{L'})}{\partial H_L(V_t, \theta)}  =  - \sum_{\phi \in \Theta} & \displaystyle \left[ \sum_{V_u, \delta(V_u, \phi)=1}{\frac{\partial \log q_{L'}(V_u, \phi)}{\partial H_L(V_t, \theta)}} \right. \\
                                                                                                                                                                 & \left. + \displaystyle  \sum_{V_u, \delta(V_u, \phi)=-1}{\frac{\partial \log\left(1 - q_{L'}(V_u, \phi) \right)}{\partial H_L(V_t, \theta)}} \right] \;.
  \end{array}
\end{equation}
Each term in this sum can be decomposed according to the chain rule of derivation, using \changed{the following equations}:
\begin{eqnarray}
  \displaystyle \frac{\partial \log \sigma(y)}{\partial \sigma(y)} &=& \frac{1}{\sigma(y)} \;,\\
  \displaystyle \frac{\partial \log\left(1 - \sigma(y) \right)}{\partial \sigma(y)} &=& \frac{-1}{1 - \sigma(y)} \;,\\
  \frac{\partial q_{L'}(V_u, \phi)}{\partial H_L(V_t, \theta)} &=& \frac{\partial q_{L'}(V_u, \phi)}{\partial \sigma(H_L(V_t, \theta))} \frac{\partial \sigma(H_L(V_t, \theta))}{\partial H_L(V_t, \theta)} \;.
  \label{eq:decompQL}
\end{eqnarray}
The second factor on the right hand side of Eq. (\ref{eq:decompQL}) can \changed{be decomposed using the derivative of the sigmoid function:}
\begin{equation}
  \label{eq:diffSigma}
  \displaystyle \frac{\partial \sigma(y)}{\partial y} = \sigma(y) (1 - \sigma(y)) \;,
\end{equation}
\added{Similarly,} the first factor on the right hand side of Eq. (\ref{eq:decompQL}) can be decomposed as follows:
\begin{equation}
  \frac{\partial q_{L'}(V_u, \phi)}{\partial p_L(V_t, \theta)} = q_{L'}(V_u, \phi)(1 - q_{L'}(V_u, \phi)) \sum_{l = 1}^{L'}{\alpha'_l \frac{\partial h'_{l}(V_u, \phi)}{\partial p_L(V_t, \theta)}} \;.
\end{equation}
\added{where $q_{L'}(V_u, \phi) = \sigma(H_{L'}(V_u, \phi))$ and $H_{L'}(V_u, \phi)$ is a function of all $p_L(V_t, \theta)$ values.}

The sample weights we have defined for CNN boosting inside a ``CNN+RNN'' network are more complex than the general case [see Eq. \ref{eq:valueOmega}]. However, they are only computed once per boosting iteration. Therefore, they do not make the optimization problem significantly less tractable, as opposed to the end-to-end training of a ``CNN+RNN'' network. But, like end-to-end training, they ensure a good complementarity between the CNN and RNN blocks.

\subsection{Joint CNN and RNN Boosting}

\changed{Two strategies are proposed below to define the order in which CNNs and RNNs are trained to design data-driven ``CNN+RNN'' architectures.}

\paragraph{``Sequential'' strategy} The most straightforward solution is to boost the CNN block while $\mathcal{L}(\boldsymbol{H}_L)$ decreases, and then to boost the RNN block while $\mathcal{L}(\boldsymbol{H'}_{L'})$ decreases. Besides the use of boosting, this is the standard approach for designing ``CNN+RNN'' networks (see section \ref{sec:DeepLearningVideoAnalysis}). However, this solution suffers from the limitation described in the previous section, namely the lack of complementarity between the CNN and RNN blocks.

\paragraph{``Joint'' strategy} To overcome this limitation, we propose to design the CNN and RNN blocks inside a single boosting loop, using a single strong learner's loss function, namely $\mathcal{L}(\boldsymbol{H}_L, \boldsymbol{H'}_{L'})$. At each boosting iteration, all CNN architectures $\boldsymbol{h} \in \mathcal{H}$ and all RNN architectures $\boldsymbol{h}' \in \mathcal{H}'$ are trained (or re-trained) and only one CNN or one RNN is added to the network: the one minimizing
\begin{equation}
  \begin{array}{c}
    \left\lbrace \mathcal{L}(\boldsymbol{H}_L + \alpha \boldsymbol{h}, \boldsymbol{H'}_{L'}) \; | \; \boldsymbol{h} \in \mathcal{H}, \alpha \geq 0 \right\rbrace \\
    \bigcup \; \left\lbrace \mathcal{L}(\boldsymbol{H}_L, \boldsymbol{H'}_{L'} + \alpha' \boldsymbol{h'}) \; | \; \boldsymbol{h}' \in \mathcal{H}', \alpha' \geq 0 \right\rbrace
  \end{array} \; .
\end{equation}
Of course, in the first boosting iteration, only CNN architectures are considered: RNNs need at least one feature extractor to operate. Eq. (\ref{eq:valueOmegaPrime}) is used to define the sample weights for CNN boosting as soon as $L' \geq 1$.

\section{\changed{Surgery Video Datasets}}
\label{sec:datasets}

\added{The proposed approach is applied to tool usage annotation in two surgical video datasets: CATARACTS and Cholec80.}

\begin{figure*}[!t]
  \begin{center}
    \includegraphics[width=0.96\textwidth]{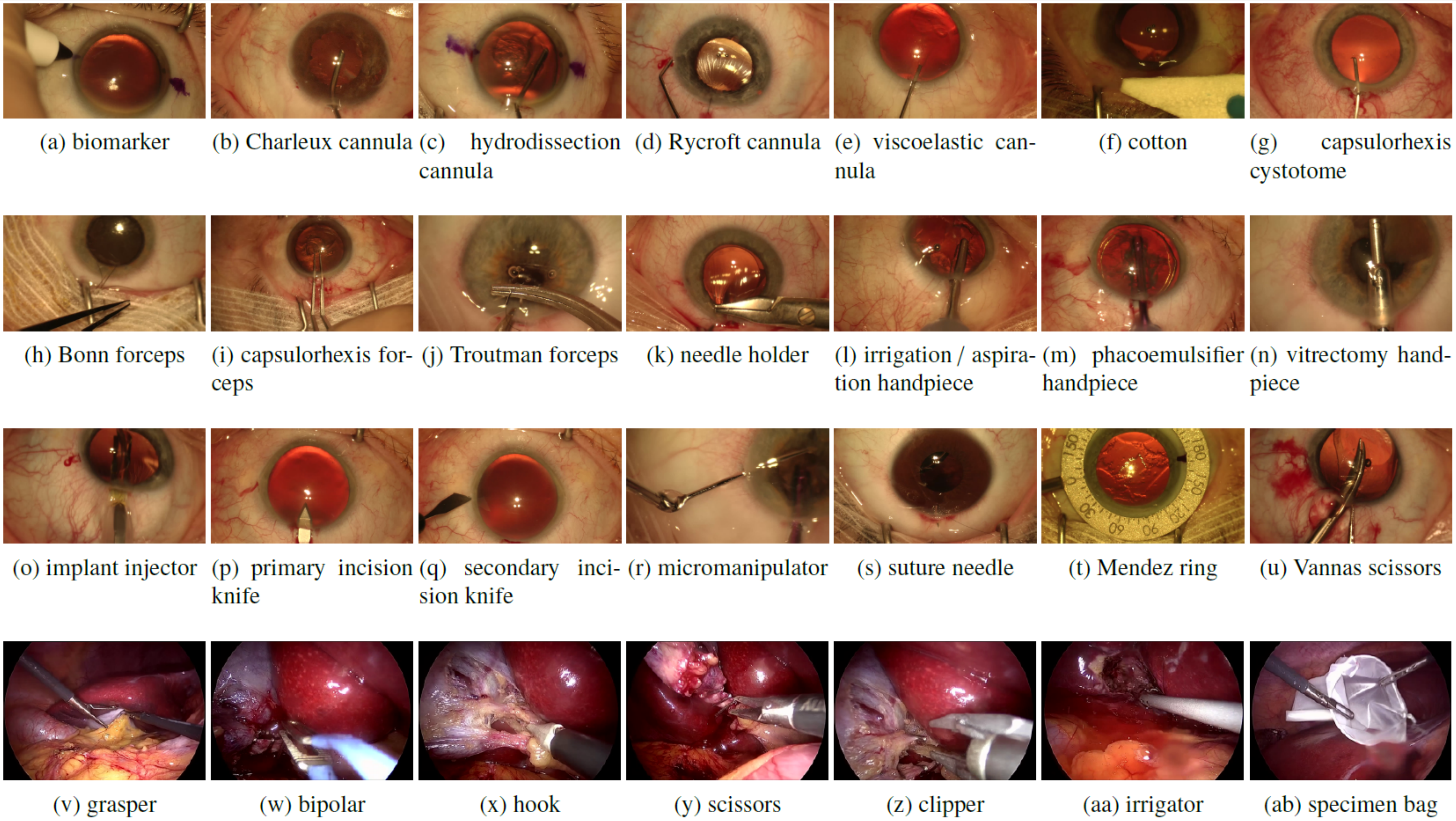}
  \end{center}
  \caption{Surgical tools annotated in videos}
  \label{fig:tools}
\end{figure*}
\begin{table}[!t]
  \caption{\changed{Statistics about tool usage annotation in the CATARACTS and Cholec80 datasets. The first column indicates inter-rater agreement (Cohen's kappa) after adjudication. The last column indicates the prevalence of each tool in the training set \changed{(excluding frames without a consensus in CATARACTS)}.}}
  \begin{center}
    \footnotesize
    \changed{\begin{tabular}{c|l|cc}
      \RotText{Dataset} & \RotText{Tool} & \RotText{Inter-rater agreement} & \RotText{\% of training frames} \\
      \hline
      \multirow{21}{*}{\RotText{CATARACTS}}
      & biomarker                       & 0.835 & 0.0168 \% \\
      & Charleux cannula                & 0.963 & 1.79 \%   \\
      & hydrodissection cannula         & 0.982 & 2.43 \%   \\
      & Rycroft cannula                 & 0.919 & 3.18 \%   \\
      & viscoelastic cannula            & 0.975 & 2.54 \%   \\
      & cotton                          & 0.947 & 0.751 \%  \\
      & capsulorhexis cystotome         & 0.995 & 4.42 \%   \\
      & Bonn forceps                    & 0.798 & 1.10 \%   \\
      & capsulorhexis forceps           & 0.849 & 1.62 \%   \\
      & Troutman forceps                & 0.764 & 0.258 \%  \\
      & needle holder                   & 0.630 & 0.0817 \% \\
      & irrigation/aspiration handpiece & 0.995 & 14.2\%    \\
      & phacoemulsifier handpiece       & 0.997 & 15.3 \%   \\
      & vitrectomy handpiece            & 0.998 & 2.76 \%   \\
      & implant injector                & 0.980 & 1.41 \%   \\
      & primary incision knife          & 0.961 & 0.700 \%  \\
      & secondary incision knife        & 0.852 & 0.522 \%  \\
      & micromanipulator                & 0.995 & 17.6 \%  \\
      & suture needle                   & 0.893 & 0.219 \%  \\
      & Mendez ring                     & 0.953 & 0.100 \%  \\
      & Vannas scissors                 & 0.823 & 0.0443 \% \\
      \hline
      \multirow{7}{*}{\RotText{Cholec80}}
      & grasper                         & n/a & 55.3 \% \\
      & bipolar                         & n/a & 4.47 \% \\
      & hook                            & n/a & 56.7 \% \\
      & scissors                        & n/a & 1.76 \% \\
      & clipper                         & n/a & 3.29 \% \\
      & irrigator                       & n/a & 5.05 \% \\
      & specimen bag                    & n/a & 6.35 \% \\
    \end{tabular}}
  \end{center}
  \label{tab:agreement}
\end{table}

\subsection{CATARACTS Dataset}
\label{sec:CATARACTSDataset}

\shrunk{The CATARACTS dataset contains 50 videos of cataract surgeries performed in Brest University Hospital.\footnote{https://cataracts.grand-challenge.org} The purpose of cataract surgeries is to remove a clouded natural lens and replace it with an artificial lens. The entire procedure can be performed with small incisions only. Surgeries were monitored through an OPMI Lumera T microscope (Carl Zeiss Meditec, Jena, Germany). Videos were recorded with a 180I camera (Toshiba, Tokyo, Japan) and a MediCap USB200 recorder (MediCapture, Plymouth Meeting, USA). The frame definition was 1920x1080 pixels and the frame rate was approximately 30 frames per second (fps). Videos had a duration of 10 minutes and 56 s on average (minimum: 6 minutes 23 s, maximum: 40 minutes 34 s). In total, more than nine hours of surgery have been video recorded. A list of 21 tools visible in these videos was compiled by a surgeon (see Fig \ref{fig:tools}). Then, the usage of each tool in videos was annotated independently by two non-clinical experts, \added{after an initial training by a surgeon.} A tool was considered to be in use whenever it was in contact with the eyeball. Therefore, both experts recorded a timestamp whenever one tool started or stopped touching the eyeball. \added{Tool-tissue contacts can be detected well: they imply deformations of the eye surface, which are well visible thanks to specular reflections of light.} Finally, annotations from both experts were adjudicated: whenever experts disagreed about the label of one tool, they watched the video together and jointly determined the actual label. However, the precise timing of tool/eyeball contacts was not adjudicated. Inter-rater agreement \changed{after adjudication} is reported in Table \ref{tab:agreement}. The dataset was divided into a training set (25 videos) and a test set (25 videos). Division was made in such a way that each tool appears in the same number of videos from both subsets (plus or minus one). The classification performance for $\theta$ was assessed only in frames where experts agreed about the usage of $\theta$. During training, some tool $\theta \in \Theta$ was considered to be in use if at least one expert said so.}

\subsection{\added{Cholec80 Dataset}}
\label{sec:Cholec80Dataset}

\added{The Cholec80 dataset contains 80 videos of cholecystectomy surgeries \citep{twinanda_endonet:_2017}. The purpose of cholecystectomy is to remove the gallbladder: this operation can be performed laparoscopically and monitored through an endoscope. Videos were recorded with a frame definition of 1920x1080 pixels and a frame rate of 25 fps. Videos had a duration of 38 minutes and 26 s on average (minimum: 12 minutes 19 s, maximum: 1 hour 39 minutes 55 s). They were downsampled to 1 fps for processing. In total, more than 51 hours of surgery have been video recorded (2 hours after downsampling). In Cholec80, a tool was considered to be in use if it was visible through the endoscope (if at least half of the tool tip was visible, precisely). The presence of seven tools was annotated in videos (see Fig \ref{fig:tools}): one binary label is provided per image and per tool. The dataset was divided into a training set (40 videos) and a test set (40 videos).}

\subsection{Training and Validation Subsets}

\changed{For validation purposes, two training videos of CATARACTS (respectively four videos of Cholec80) were assigned to a validation subset; the remaining training videos were assigned to a learning subset used to optimize the CNN, RNN and boosting weights. In CATARACTS,} the validation videos were chosen such that all tools appear in the learning subset: it was not possible to ensure this property for both subsets. \added{In Cholec80, they were chosen at random.}

\section{Experiments}
\label{sec:Experiments}

\subsection{Architectures}

\changed{Seven CNN architectures were used as weak classifiers in this paper:
\begin{itemize}
  \item VGG-16 and VGG-19 \citep{simonyan_very_2015},
  \item the second version \citep{he_identity_2016} of ResNet-101 and ResNet-152 \citep{he_deep_2016},
  \item Inception-v4 and Inception-ResNet-v2 \citep{szegedy_inception-v4_2017},
  \item NASNet-A \citep{zoph_learning_2017}.
\end{itemize}
The TensorFlow-Slim implementation\footnote{\url{https://github.com/tensorflow/models/tree/master/research/slim}} of these CNNs was used, with weights pre-trained on ImageNet. The last layer of each CNN, which computes one logit prediction per class, was resized from 1000 neurons for ImageNet to 21 neurons for CATARACTS or 7 neurons for Cholec80; the weights of these neurons were initialized at random. The same input image size was used for ImageNet, CATARACTS and Cholec80: $224 \times 224$ pixels for VGG-16 and VGG-19, $299 \times 299$ pixels for ResNet-101, ResNet-152, Inception-v4 and Inception-ResNet-v2, and $331\times 331$ pixels for NASNet-A. To preserve the aspect ratio, images from CATARACTS and Cholec80 were first resized to $224 \times 126$ pixels, $299 \times 168$ pixels or $331\times 184$ pixels and were then padded with zeros at the top and the bottom to obtain square images.} \added{All CNNs were trained using the RMSProp algorithm with a learning rate initialized to $0.01$ and decaying exponentially. In order to define a more challenging boosting problem, we conducted a secondary experiment involving the three worst performing CNNs only: this experiment is called ``weaker CNNs'', while the primary experiment involving all CNNs is called ``all CNNs''.}

\begin{figure}[!t]
  \begin{center}
    \includegraphics[width=.8\textwidth]{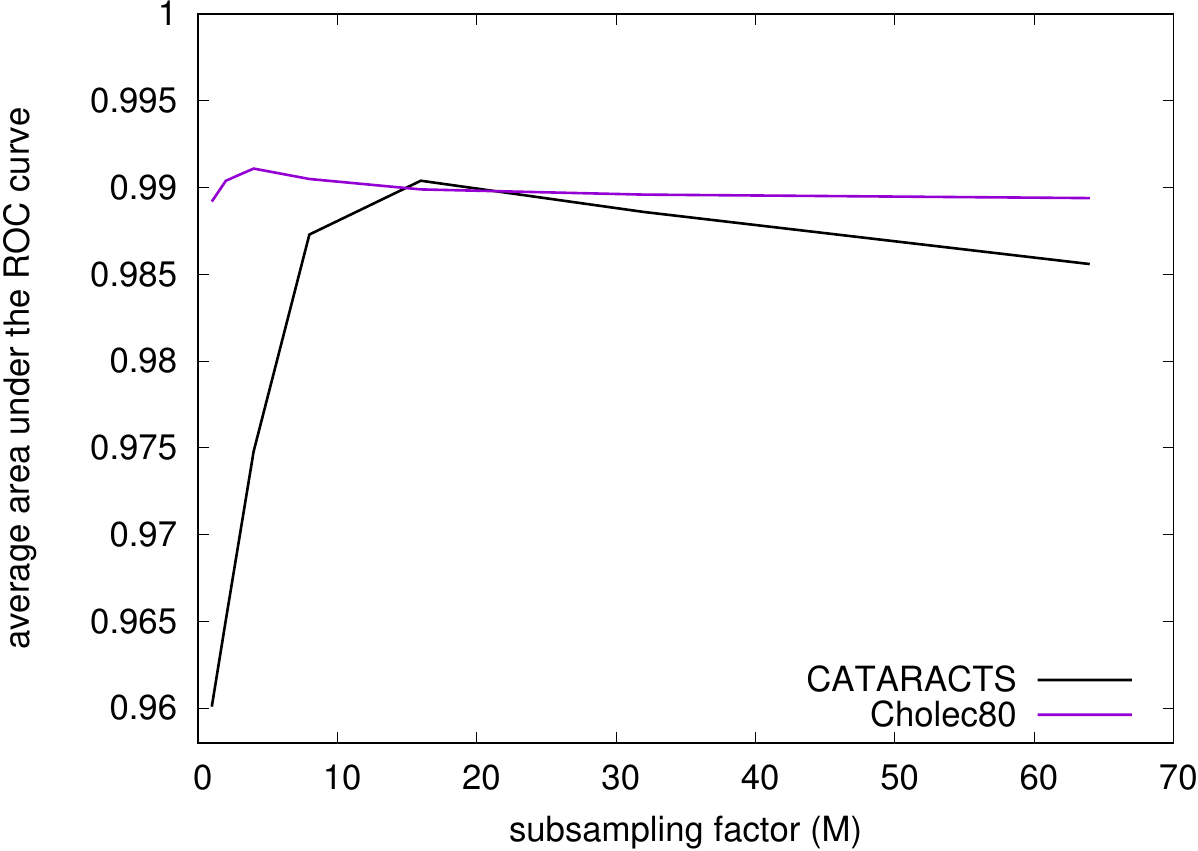}
  \end{center}
  \caption{\changed{Effect of the subsampling factor $M$ (which is also the data augmentation rate --- see section \ref{sec:RNNsLongVideoSequences}) on tool annotation performance in the validation subset. This figure reports the average performance obtained using NASNet-A and each of the six weak RNN classifiers based on LSTM or GRU.}}
  \label{fig:subsamplingRate}
\end{figure}

Regarding RNN boosting, two types of RNN cells were used: LSTM \citep{hochreiter_long_1997} and GRU \citep{cho_properties_2014}. To limit complexity and computation times, the number of layers in RNNs was set to $n = 2$. Three different values were used for $C$, the number of neurons per cell, in order to define six weak classifiers (three based on LSTM, three based on GRU): $C = 64$, $C = 128$, $C = 256$. In all RNN boosting experiments, a subsampling factor of \changed{$M = 16$ and $M = 4$ was used in CATARACTS and Cholec80}, respectively: this number was found to be optimal in initial experiments on the validation subset (see Fig. \ref{fig:subsamplingRate}). \added{All RNNs were trained using the RMSProp algorithm with a constant learning rate of $0.001$. As for the median filter radii $R_\theta$, they were selected within $\{1, 2, 4, 8, 16, 32, 64\}$ to maximize the classification performance in the validation set; for rare tools absent from the validation set, the most frequently selected value was used.}
RNNs were implemented using Keras version 2.0.8.

Inference times for CNNs, the most computationally intensive parts of the system, are reported in Table \ref{tab:computationTimes}.
\begin{table}[!t]
  \caption{Inference times of CNNs using one GeForce GTX \changed{1080 Ti} GPU by Nvidia. Inference times are given for batch processing \changed{(mini-batches of 16 images for NASNet-A and 32 images for other CNNs)}, which can be used for offline video labeling, and for single image processing, which must be used for online video labeling.}
  \changed{\begin{center}
    \footnotesize
    \begin{tabular}{c|c|c}
    CNN                 & single image & batch processing \\
    \hline
    VGG-16              & 7.50 ms / image  & 2.87 ms / image \\
    VGG-19              & 8.50 ms / image  & 3.44 ms / image \\
    ResNet-101          & 10.2 ms / image  & 3.16 ms / image \\
    ResNet-152          & 13.2 ms / image  & 4.62 ms / image \\
    Inception-v4        & 18.8 ms / image  & 6.09 ms / image \\
    Inception-ResNet-v2 & 19.0 ms / image  & 6.34 ms / image \\
    NASNet-A            & 24.6 ms / image  & 18.5 ms / image \\
    \end{tabular}
  \end{center}}
  \label{tab:computationTimes}
\end{table}

\subsection{Performance of Boosted Video Labelers}

\begin{figure*}[!t]
  \begin{center}
    \changed{\begin{tabular}{cc}
      \subfloat[CATARACTS]{\includegraphics[width=0.475\textwidth]{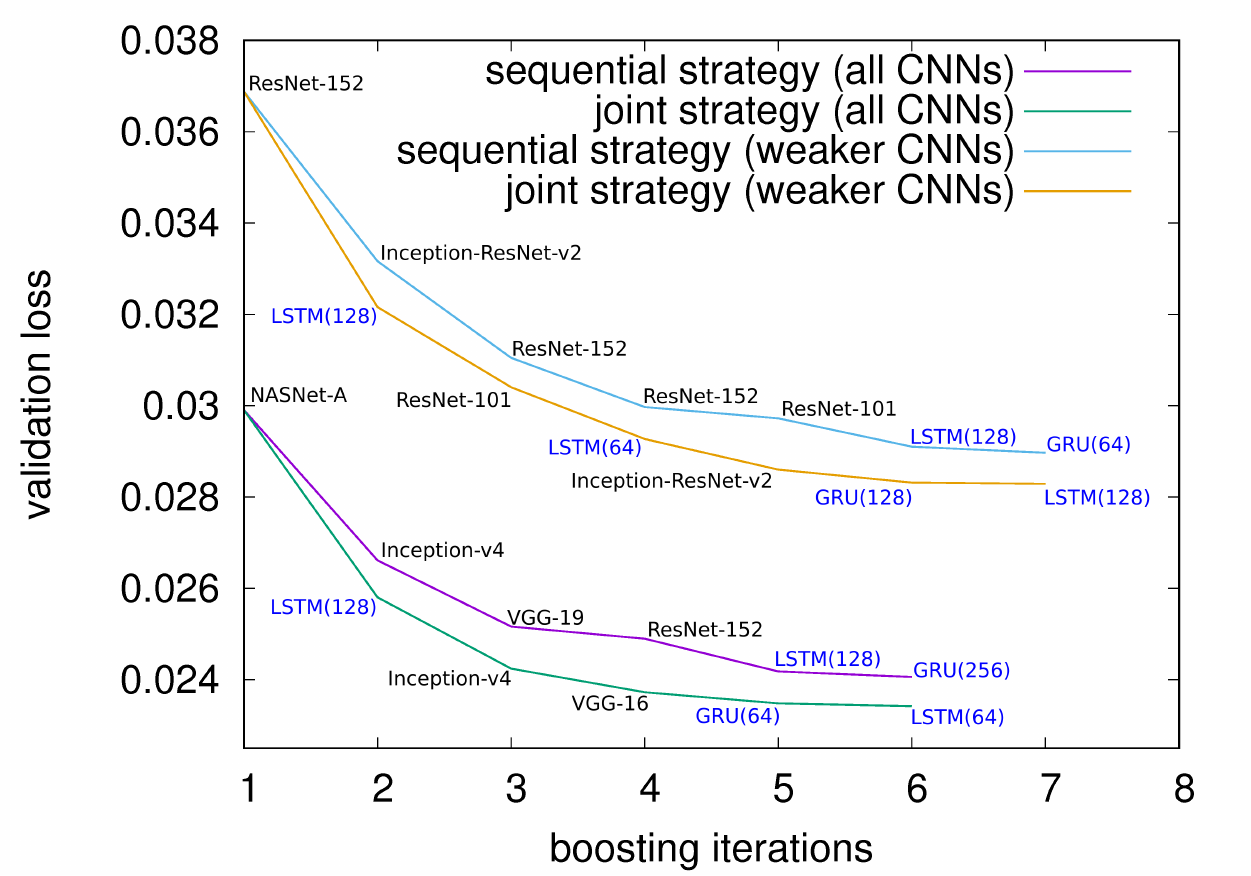}} &
      \subfloat[Cholec80]{\includegraphics[width=0.475\textwidth]{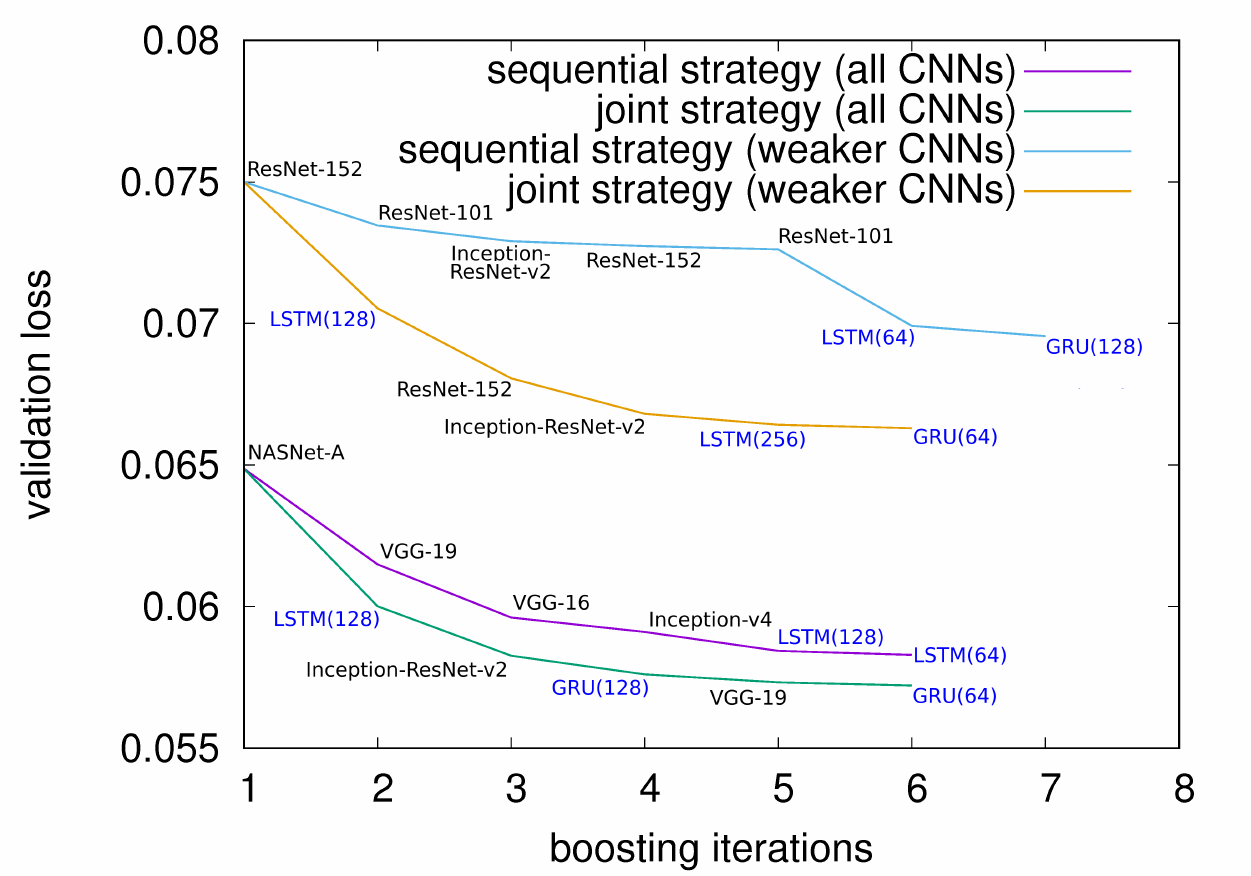}}
    \end{tabular}}
  \end{center}
  \caption{\added{Evolution of the validation loss across boosting iterations, using bidirectional RNNs.} The number of neurons in RNN cells is indicated in brackets. \added{Curves and architectures obtained for the unidirectional version are very similar: they are not reported.}}
  \label{fig:boosting}
\end{figure*}

\added{The performance of the seven weak CNNs is reported in Tables \ref{tab:Az} and \ref{tab:AP}. As expected, the best performing CNN, NASNet-A, is also the most recent. Surprisingly, VGG-19 and VGG-16 are also quite good, in spite of being older and less sophisticated than the others. The three worst performing CNNs (in the validation set and in the test set) are ResNet-101, ResNet-152 and Inception-ResNet-v2: they were used in the ``weaker CNNs'' experiment.} \changed{The architecture of boosted bidirectional video labelers are reported in Fig \ref{fig:boosting} for the ``all CNNs'' and ``weaker CNNs'' experiments.} \added{Their performance is detailed in Tables \ref{tab:Az} and \ref{tab:AP} for the ``all CNNs'' experiment. In the largest dataset (CATARACTS), training the initial CNNs with early stopping took between 2h (ResNet-101) and 11h (Inception-ResNet-v2); training NASNet-A took 8h. In the following boosting iterations, fine-tuning the CNNs and training/fine-tuning the RNNs took 3h at most per CNN or RNN. At each boosting iterations, CNNs and RNNs were trained in parallel on a cluster of GeForce GTX 1080 Ti GPUs (RNNs were trained without GPU). Overall, if the process was fully-automated, boosting would have lasted approximately 29h. In practice, it took a few days, as the process involved manual interactions (for early stopping in particular). The end-to-end training of a NASNet-A + RNN network would have lasted more than 80,000 hours (9 years) for CATARACTS, which involves sequences of more than 10,000 frames.}

\begin{figure*}[!t]
  \begin{center}
    \changed{\begin{tabular}{ccc}
      \subfloat[NASNet-A -- CATARACTS]{\includegraphics[width=0.3\textwidth]{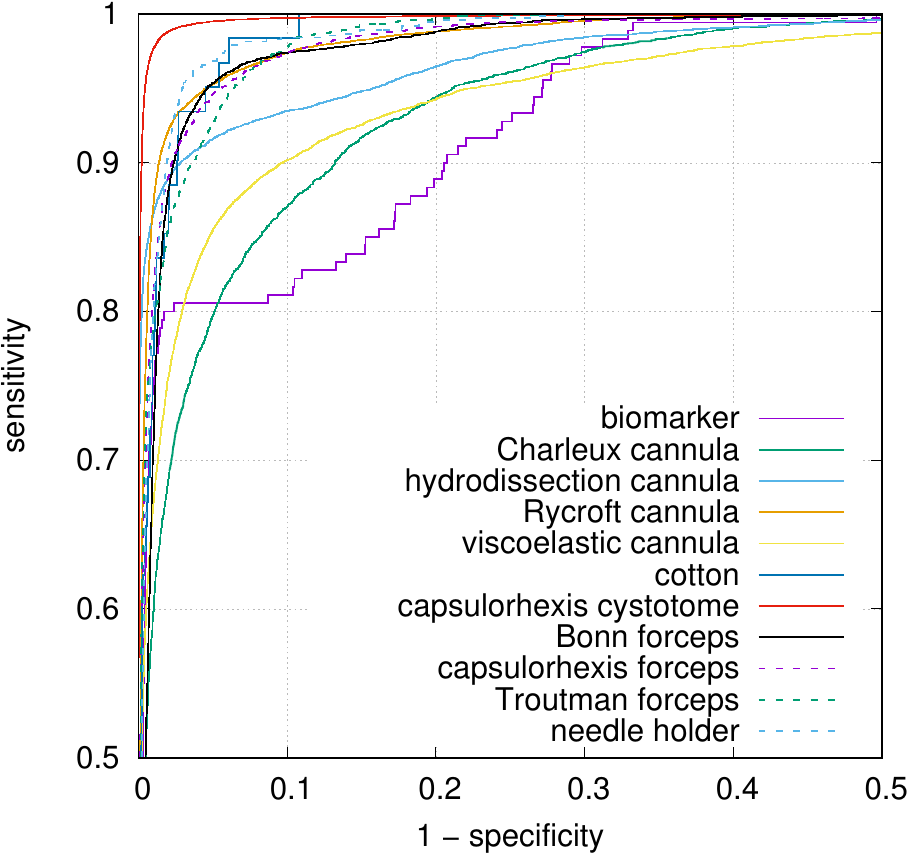}} &
      \subfloat[NASNet-A -- CATARACTS]{\includegraphics[width=0.3\textwidth]{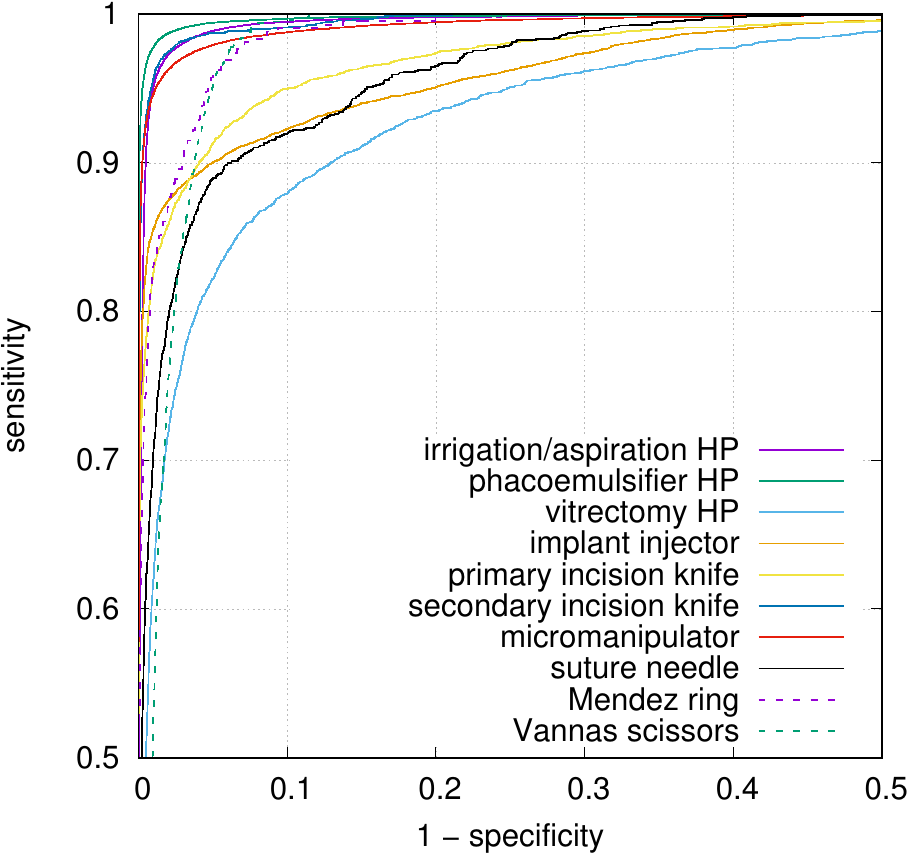}} &
      \subfloat[NASNet-A -- Cholec80]{\includegraphics[width=0.3\textwidth]{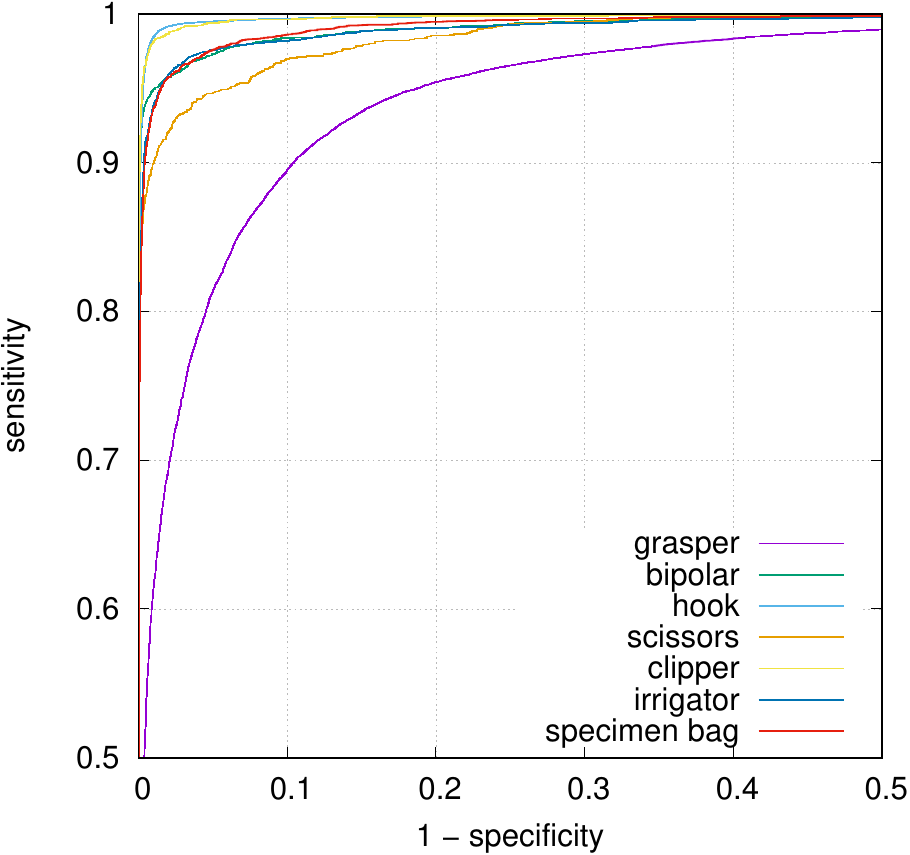}} \\
      \subfloat[best ensemble -- CATARACTS]{\includegraphics[width=0.3\textwidth]{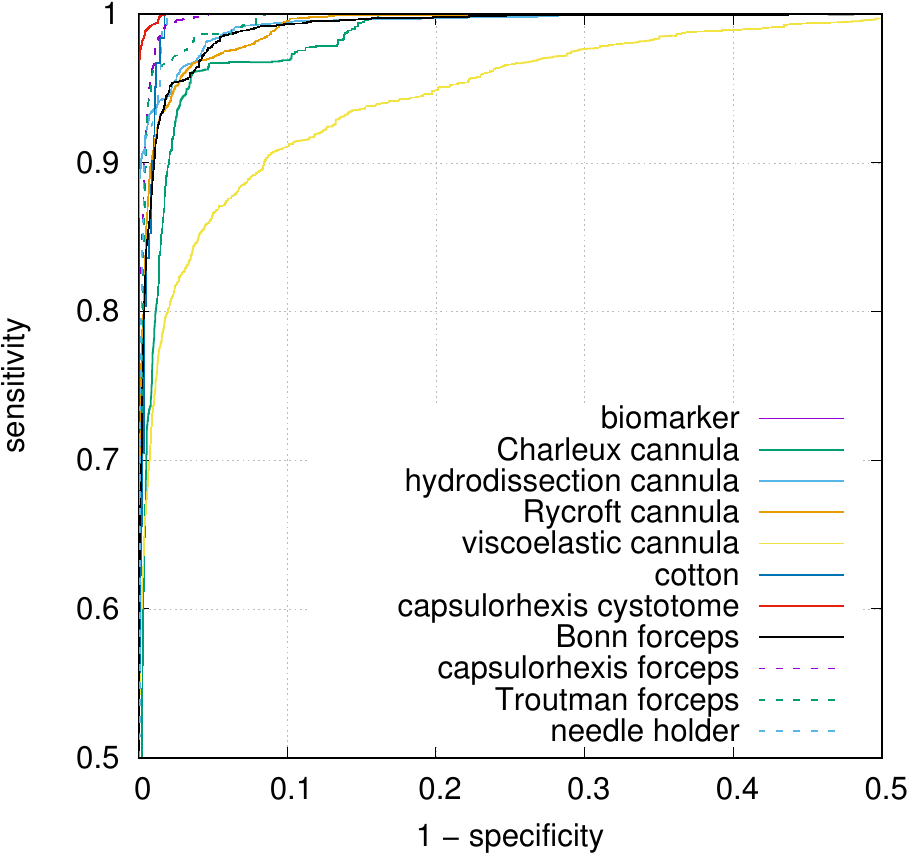}} &
      \subfloat[best ensemble -- CATARACTS]{\includegraphics[width=0.3\textwidth]{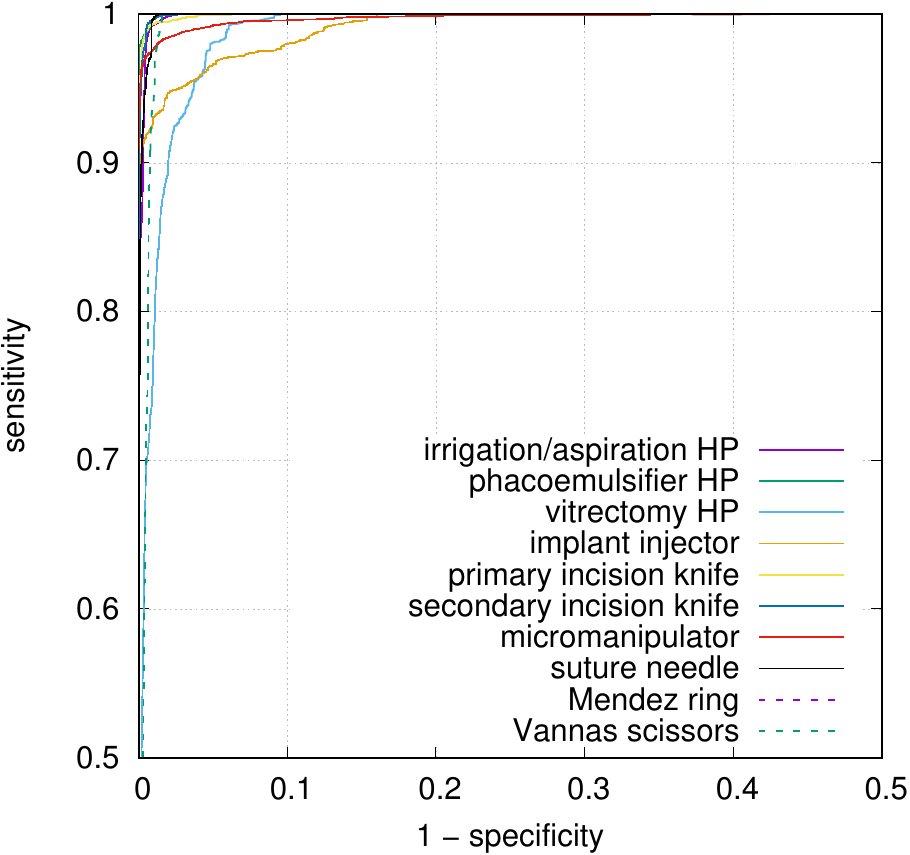}} &
      \subfloat[best ensemble -- Cholec80]{\includegraphics[width=0.3\textwidth]{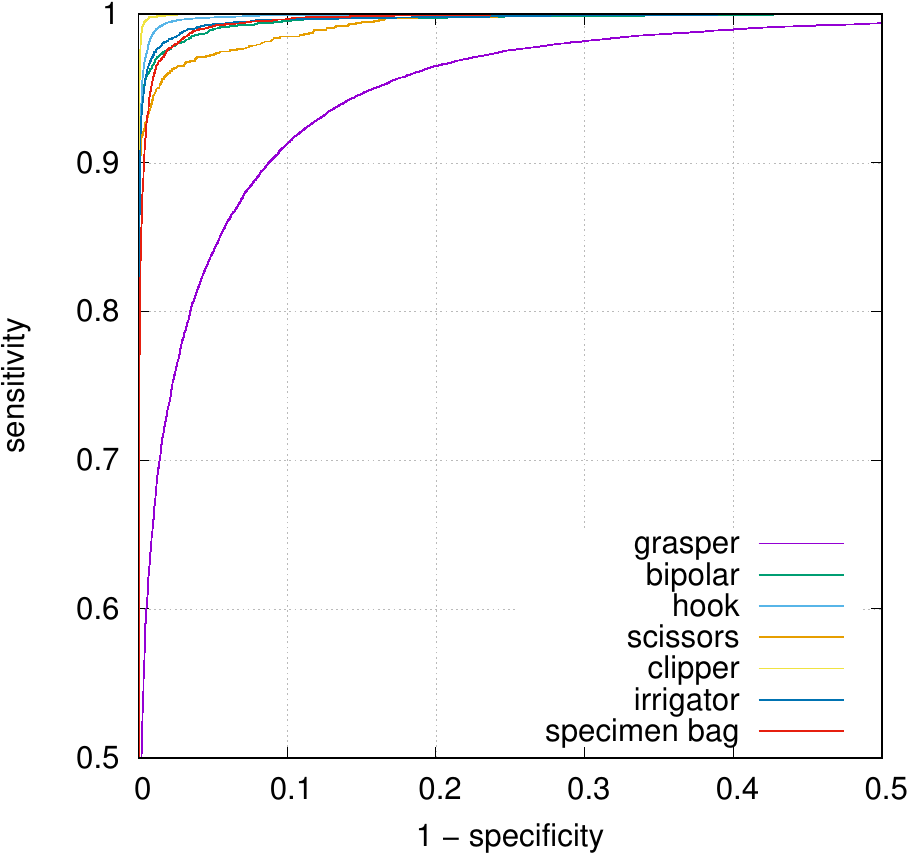}} \\
    \end{tabular}}
  \end{center}
  \caption{\changed{Receiver-operating characteristic (ROC) curves for the best weak classifier (NASNet-A) and the best ensemble of the ``all CNNs'' experiment (jointly boosted ``CNN+RNN'' architecture, with median filtering for CATARACTS). Note that only the top left quadrant of the ROC space (sensitivity and specificity $\geq$ 0.5) is displayed for improved visualization.}}
  \label{fig:ROC}
\end{figure*}

\begin{figure*}[!t]
  \begin{center}
    \added{\begin{tabular}{ccc}
      \subfloat[NASNet-A -- CATARACTS]{\includegraphics[width=0.3\textwidth]{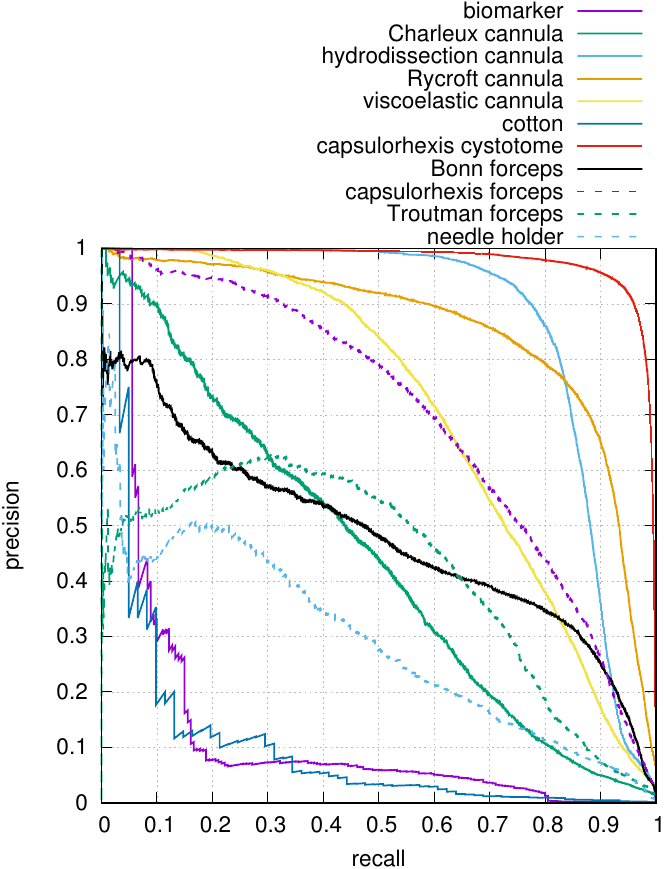}} &
      \subfloat[NASNet-A -- CATARACTS]{\includegraphics[width=0.3\textwidth]{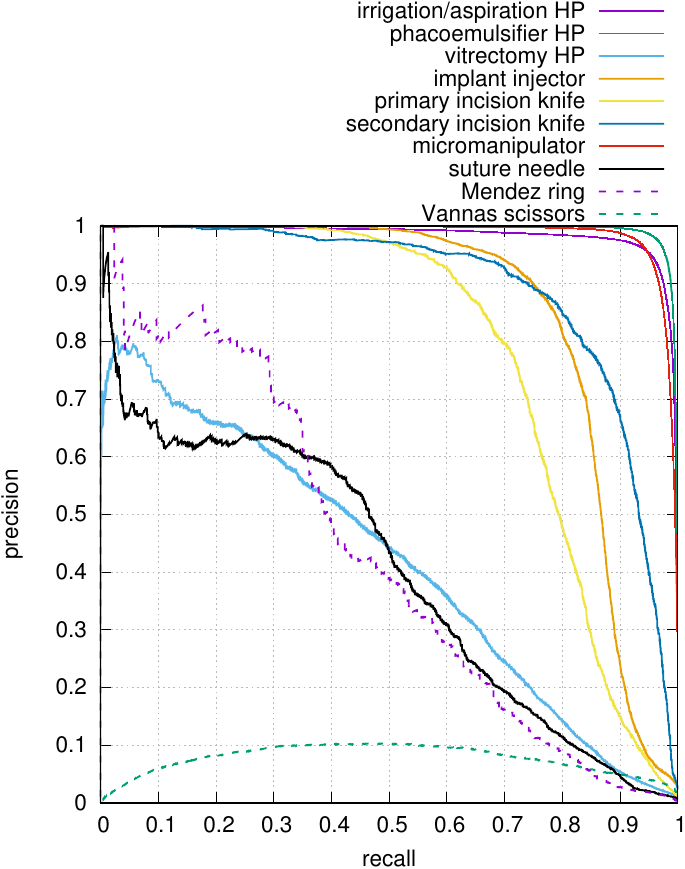}} &
      \subfloat[NASNet-A -- Cholec80]{\includegraphics[width=0.3\textwidth]{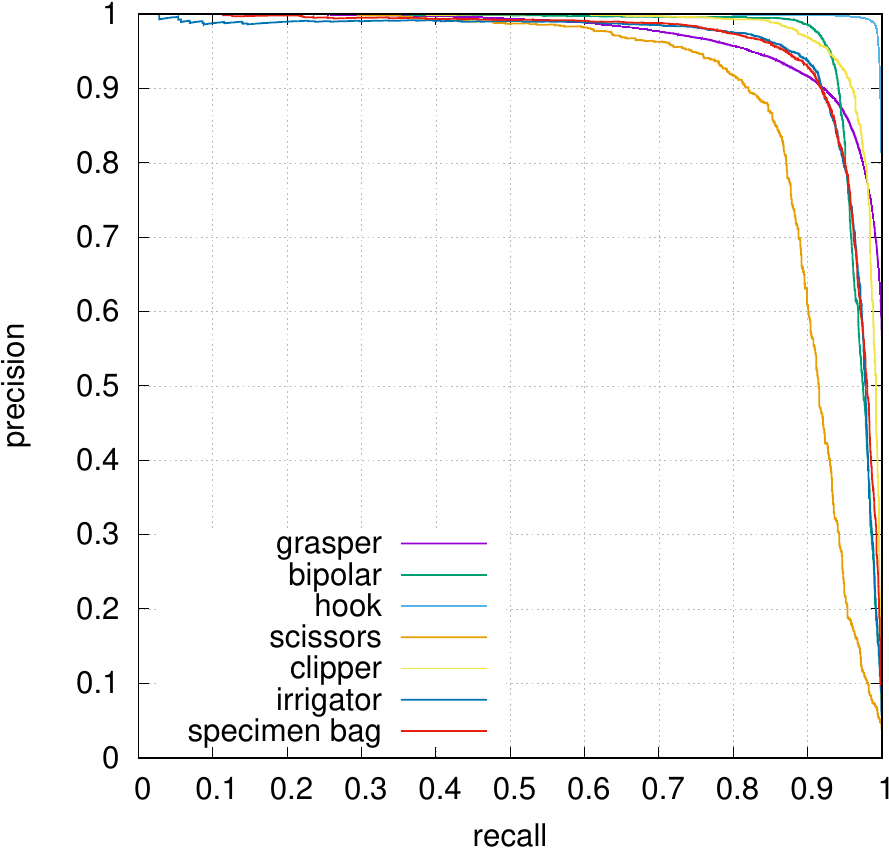}} \\
      \subfloat[best ensemble -- CATARACTS]{\includegraphics[width=0.3\textwidth]{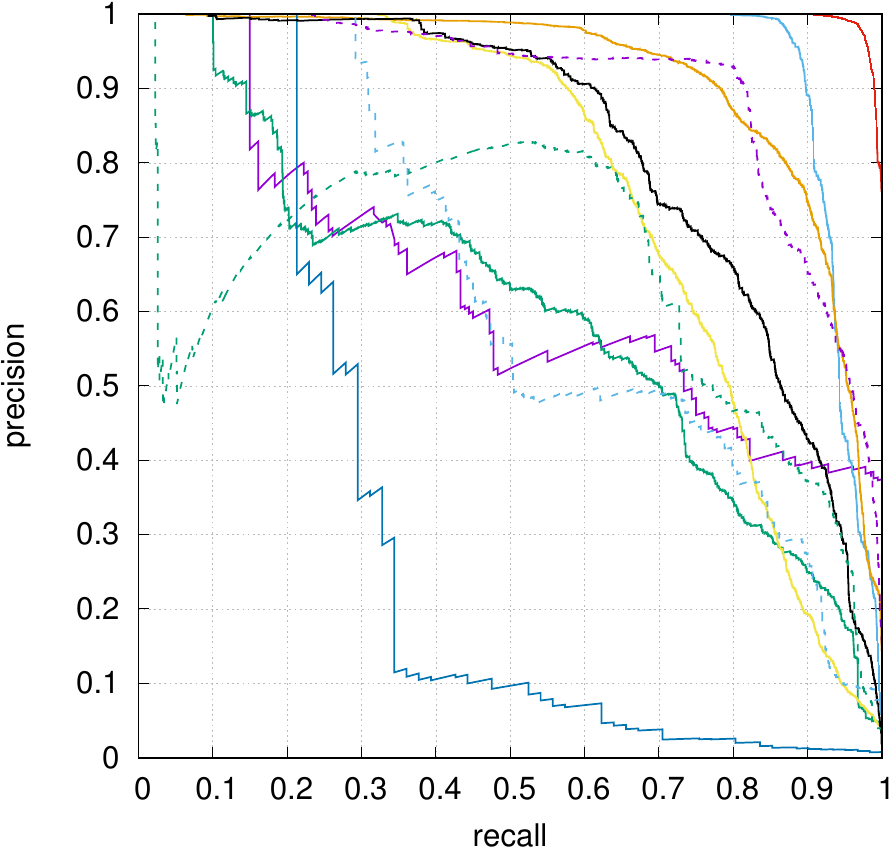}} &
      \subfloat[best ensemble -- CATARACTS]{\includegraphics[width=0.3\textwidth]{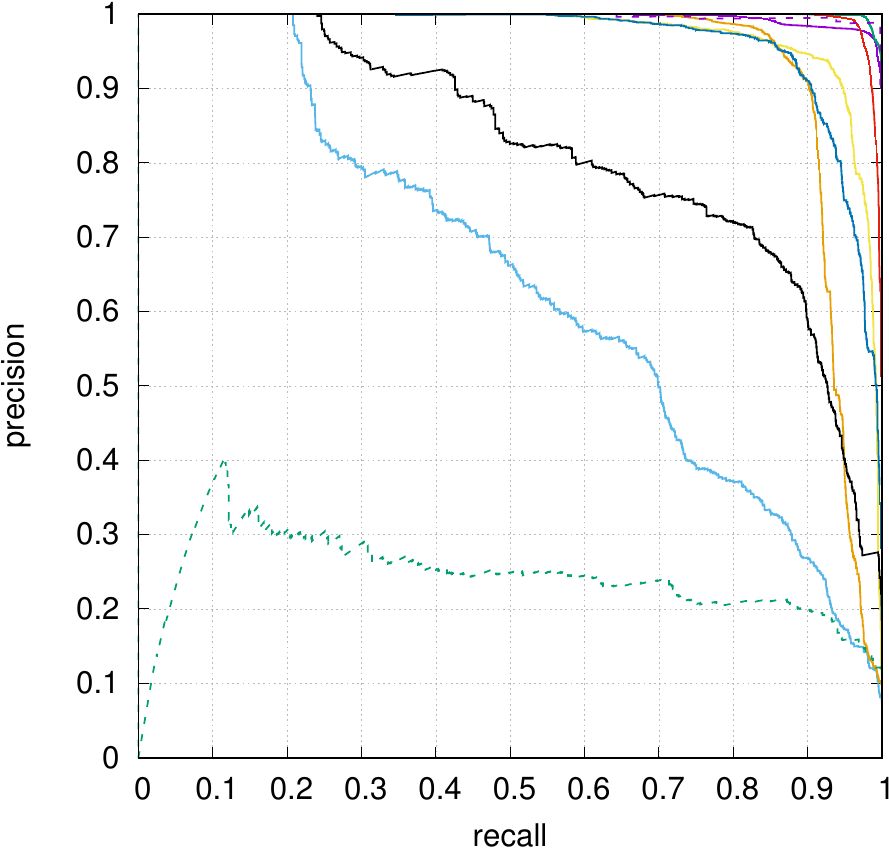}} &
      \subfloat[best ensemble -- Cholec80]{\includegraphics[width=0.3\textwidth]{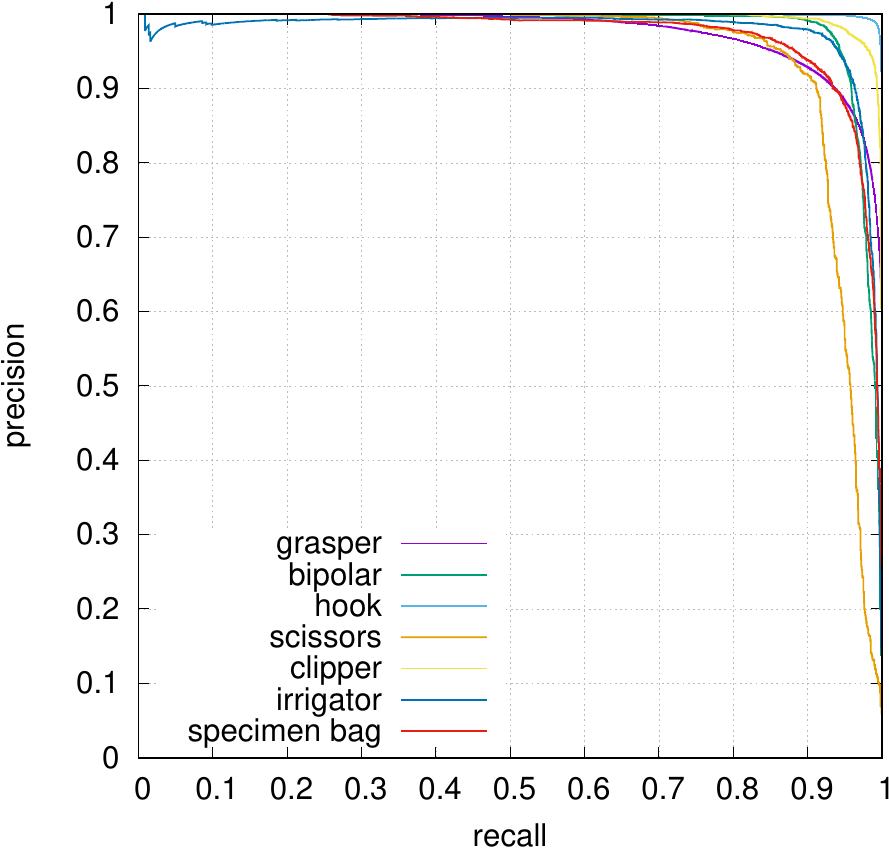}} \\
    \end{tabular}}
  \end{center}
  \caption{\added{Precision-recall (PR) curves for the best weak classifier (NASNet-A) and the best ensemble of the ``all CNNs'' experiment (jointly boosted ``CNN+RNN'' architecture, with median filtering for CATARACTS). Fig. (a) and (d) share the same legend. The same applies to Fig. (b) and (e).}}
  \label{fig:PR}
\end{figure*}

\added{Tables \ref{tab:Az} and \ref{tab:AP} show that ``CNN+RNN'' boosting improves performance compared to CNN boosting alone in both datasets. Median filtering also improves performance in the CATARACTS dataset but decreases it in Cholec80. For each tool $\theta$, the least worst radius is $R_\theta = 1$ for Cholec80 and the best radius is $2 \leq R_\theta \leq 32$ for CATARACTS. ROC curves and precision-recall curves for the best CNN, namely NASNet-A, and the best ensemble, namely joint ``CNN+RNN'' boosting (with median filtering for CATARACTS), are reported in Fig. \ref{fig:ROC} and \ref{fig:PR}.} \added{In terms of area under the ROC curve ($A_z$), all tools were detected well by the best ensemble ($A_z \geq 0.9694$). In terms of average precision (AP), rare tools are poorly detected before boosting ($AP < 0.1$ in some cases). For rare tools, precision (and therefore AP) is indeed impacted strongly by the number of false alarms which, in the specificity criterion (and therefore $A_z$), is divided by the large number of negative samples. In fact, as shown in Table \ref{tab:AP}, AP is highly correlated with tool prevalence in the training set. However, the mean AP is greatly improved after boosting: from mAP $= 0.6086$ to mAP $= 0.7980$ in CATARACTS. Since there are no rare tools in Cholec80, mAP is much higher (up to mAP $= 0.9789$).}

\begin{table*}[!t]
  \caption{\changed{Areas under the ROC curves ($A_z$) for each weak CNN classifier and strong classifiers in the ``all CNNs'' experiment. In case of ``CNN+RNN'' boosting, the ``joint'' strategy is used. HP stands for ``handpiece''. On each line, the highest score is marked in bold and the highest score among weak CNN classifiers is marked in italic. For each dataset, the last row indicates the Pearson correlation between $A_z$ in the test set and tool prevalence in the training set (see Table \ref{tab:agreement}).}}
  \begin{center}
    \footnotesize
    \changed{\begin{tabular}{c|l|ccccccc|ccc}
      \RotText{Dataset} & \RotText{Tool} & \RotText{VGG-16} & \RotText{VGG-19} & \RotText{ResNet-101} & \RotText{ResNet-152} & \RotText{Inception-v4} & \RotText{Inception-ResNet-v2} & \RotText{NASNet-A} & \RotText{boosted CNN} & \RotText{boosted CNN+RNN} & \RotText{smoothed boosted CNN+RNN} \\
      \hline
      \multirow{22}{*}{\RotText{CATARACTS}}
      & biomarker                & 0.9364 & 0.9855 & 0.9469 & \emph{0.9948} & 0.7852 & 0.6313 & 0.9544 & 0.9920 & 0.9997 & \textbf{0.9998} \\
      & Charleux cannula         & 0.9105 & 0.9360 & 0.8238 & 0.8813 & 0.9129 & 0.9174 & \emph{0.9603} & 0.9677 & 0.9877 & \textbf{0.9903} \\
      & hydrodissection cannula  & 0.9807 & 0.9875 & 0.9585 & 0.9709 & \emph{0.9911} & 0.9768 & 0.9807 & 0.9938 & 0.9956 & \textbf{0.9965} \\
      & Rycroft cannula          & 0.9858 & 0.9846 & 0.9747 & 0.9776 & 0.9857 & 0.9743 & \emph{0.9896} & 0.9920 & 0.9948 & \textbf{0.9952} \\
      & viscoelastic cannula     & 0.9441 & 0.9341 & 0.9347 & 0.8709 & 0.9393 & 0.9224 & \emph{0.9628} & 0.9630 & 0.9670 & \textbf{0.9711} \\
      & cotton                   & 0.9879 & 0.9889 & 0.9417 & 0.9848 & 0.9622 & 0.9472 & \emph{0.9910} & 0.9926 & 0.9966 & \textbf{0.9973} \\
      & capsulorhexis cystotome  & 0.9935 & 0.9978 & 0.9950 & 0.9955 & 0.9981 & 0.9940 & \emph{0.9982} & 0.9990 & 0.9997 & \textbf{0.9998} \\
      & Bonn forceps             & 0.9769 & \emph{0.9896} & 0.9745 & 0.9813 & 0.9867 & 0.9768 & 0.9860 & 0.9933 & 0.9945 & \textbf{0.9949} \\
      & capsulorhexis forceps    & 0.9706 & 0.9767 & 0.9648 & 0.9641 & \emph{0.9896} & 0.9765 & 0.9878 & 0.9944 & 0.9983 & \textbf{0.9986} \\
      & Troutman forceps         & 0.9746 & 0.9811 & 0.9790 & 0.9472 & 0.9844 & 0.9766 & \emph{0.9886} & 0.9901 & 0.9969 & \textbf{0.9971} \\
      & needle holder            & 0.9667 & \emph{0.9911} & 0.9329 & 0.9312 & 0.9722 & 0.9762 & 0.9908 & 0.9951 & 0.9974 & \textbf{0.9981} \\
      & irrigation/aspiration HP & 0.9950 & 0.9960 & 0.9879 & 0.9910 & 0.9961 & 0.9913 & \emph{0.9963} & 0.9981 & 0.9991 & \textbf{0.9993} \\
      & phacoemulsifier HP       & 0.9969 & 0.9983 & 0.9939 & 0.9968 & 0.9980 & 0.9969 & \emph{0.9984} & 0.9992 & \textbf{0.9998} & \textbf{0.9998} \\
      & vitrectomy HP            & 0.9756 & 0.9761 & 0.9874 & 0.9516 & 0.9812 & \emph{0.9888} & 0.9579 & 0.9930 & 0.9894 & \textbf{0.9927} \\
      & implant injector         & 0.9811 & 0.9827 & 0.9790 & 0.9772 & \emph{0.9887} & 0.9797 & 0.9765 & 0.9909 & 0.9943 & \textbf{0.9952} \\
      & primary incision knife   & 0.9881 & 0.9908 & 0.9819 & 0.9686 & \emph{0.9959} & 0.9809 & 0.9814 & 0.9969 & 0.9994 & \textbf{0.9996} \\
      & secondary incision knife & 0.9924 & 0.9976 & 0.9977 & \emph{0.9989} & 0.9982 & 0.9980 & 0.9972 & 0.9989 & 0.9996 & \textbf{0.9997} \\
      & micromanipulator         & 0.9919 & 0.9943 & 0.9919 & 0.9913 & \emph{0.9959} & 0.9922 & 0.9953 & 0.9972 & 0.9983 & \textbf{0.9985} \\
      & suture needle            & 0.9757 & \emph{0.9779} & 0.9504 & 0.9742 & 0.9647 & 0.9612 & 0.9756 & 0.9844 & 0.9989 & \textbf{0.9991} \\
      & Mendez ring              & 0.9943 & \emph{0.9997} & 0.9939 & 0.9792 & 0.9435 & 0.9724 & 0.9913 & 0.9970 & \textbf{1.0000} & \textbf{1.0000} \\
      & Vannas scissors          & \emph{0.9939} & 0.9924 & 0.9810 & 0.9648 & 0.9799 & 0.9662 & 0.9842 & 0.9943 & 0.9947 & \textbf{0.9961} \\
      \cline{2-12}
      & Average (m$A_z$)         & 0.9768 & \emph{0.9837} & 0.9653 & 0.9663 & 0.9690 & 0.9570 & 0.9831 & 0.9916 & 0.9953 & \textbf{0.9961} \\
      \cline{2-12}
      & Corr. with prevalence    & 0.3246 & 0.2305 & 0.2793 & 0.2850 & 0.2955 & 0.2461 & 0.3922 & 0.2493 & 0.1309 & 0.1220 \\
      \hline
      \multirow{8}{*}{\RotText{Cholec80}}
      & grasper                  & \emph{0.9633} & 0.9620 & 0.9472 & 0.9539 & 0.9505 & 0.9523 & 0.9618 & 0.9689 & \textbf{0.9694} & 0.9652 \\
      & bipolar                  & \emph{0.9949} & 0.9946 & 0.9929 & 0.9904 & 0.9903 & 0.9913 & 0.9941 & 0.9962 & \textbf{0.9977} & 0.9975 \\
      & hook                     & 0.9983 & 0.9983 & 0.9972 & 0.9975 & 0.9963 & 0.9973 & \emph{0.9984} & 0.9989 & \textbf{0.9991} & 0.9966 \\
      & scissors                 & 0.9877 & 0.9865 & 0.9771 & 0.9751 & 0.9802 & 0.9819 & \emph{0.9903} & 0.9909 & 0.9958 & \textbf{0.9959} \\
      & clipper                  & 0.9977 & 0.9979 & 0.9955 & 0.9958 & 0.9923 & 0.9954 & \emph{0.9983} & 0.9989 & \textbf{0.9998} & \textbf{0.9998} \\
      & irrigator                & 0.9932 & \emph{0.9935} & 0.9859 & 0.9895 & 0.9861 & 0.9882 & 0.9930 & 0.9956 & \textbf{0.9980} & 0.9975 \\
      & specimen bag             & \emph{0.9951} & \emph{0.9951} & 0.9915 & 0.9914 & 0.9926 & 0.9937 & 0.9944 & 0.9969 & 0.9976 & \textbf{0.9977} \\
      \cline{2-12}
      & Average (m$A_z$)         & \emph{0.9900} & 0.9897 & 0.9839 & 0.9848 & 0.9840 & 0.9857 & \emph{0.9900} & 0.9923 & \textbf{0.9939} & 0.9929 \\
      \cline{2-12}
      & Corr. with prevalence    & -0.4916 & -0.4865 & -0.4317 & -0.3751 & -0.4396 & -0.4555 & -0.5149 & -0.5186 & -0.5915 & -0.6535 \\
    \end{tabular}}
  \end{center}
  \label{tab:Az}
\end{table*}

\begin{table*}[!t]
  \caption{\added{Average precision (AP) for each weak CNN classifier and strong classifiers in the ``all CNNs'' experiment. In case of ``CNN+RNN'' boosting, the ``joint'' strategy is used. HP stands for ``handpiece''. On each line, the highest score is marked in bold and the highest score among weak CNN classifiers is marked in italic. For each dataset, the last row indicates the Pearson correlation between AP in the test set and tool prevalence in the training set (see Table \ref{tab:agreement}).}}
  \begin{center}
    \footnotesize
    \added{\begin{tabular}{c|l|ccccccc|ccc}
      \RotText{Dataset} & \RotText{Tool} & \RotText{VGG-16} & \RotText{VGG-19} & \RotText{ResNet-101} & \RotText{ResNet-152} & \RotText{Inception-v4} & \RotText{Inception-ResNet-v2} & \RotText{NASNet-A} & \RotText{boosted CNN} & \RotText{boosted CNN+RNN} & \RotText{smoothed boosted CNN+RNN} \\
      \hline
      \multirow{22}{*}{\RotText{CATARACTS}}
      & biomarker                & 0.0046 & 0.0120 & 0.0039 & 0.0482 & 0.0012 & 0.0005 & \emph{0.1294} & 0.1311 & 0.5628 & \textbf{0.6352} \\
      & Charleux cannula         & 0.0538 & 0.0891 & 0.0473 & 0.1353 & 0.1276 & 0.1594 & \emph{0.4386} & 0.2455 & 0.5728 & \textbf{0.6003} \\
      & hydrodissection cannula  & 0.8339 & 0.8652 & 0.7870 & 0.8211 & \emph{0.8881} & 0.8141 & 0.8678 & 0.9213 & 0.9412 & \textbf{0.9471} \\
      & Rycroft cannula          & 0.7819 & 0.7095 & 0.7381 & 0.7357 & 0.8085 & 0.7807 & \emph{0.8530} & 0.8637 & 0.9084 & \textbf{0.9155} \\
      & viscoelastic cannula     & 0.5670 & 0.6065 & 0.5925 & 0.5582 & 0.6178 & 0.4828 & \emph{0.7048} & 0.6833 & 0.7588 & \textbf{0.7658} \\
      & cotton                   & 0.0063 & 0.0071 & 0.0101 & 0.1317 & 0.1093 & \emph{0.2491} & 0.1092 & 0.1474 & 0.2308 & \textbf{0.3148} \\
      & capsulorhexis cystotome  & 0.9356 & 0.9700 & 0.9399 & 0.9510 & 0.9768 & 0.9462 & \emph{0.9786} & 0.9868 & 0.9959 & \textbf{0.9968} \\
      & Bonn forceps             & 0.6181 & 0.7007 & 0.6580 & 0.7102 & \emph{0.7251} & 0.5806 & 0.4816 & 0.7805 & 0.8174 & \textbf{0.8223} \\
      & capsulorhexis forceps    & 0.6319 & 0.6441 & 0.6343 & 0.6399 & \emph{0.7705} & 0.6600 & 0.6956 & 0.8210 & 0.8950 & \textbf{0.9023} \\
      & Troutman forceps         & 0.2468 & 0.2408 & 0.3803 & 0.2779 & 0.3714 & 0.3282 & \emph{0.4200} & 0.4348 & 0.6173 & \textbf{0.6474} \\
      & needle holder            & 0.1371 & 0.2420 & 0.0495 & 0.0514 & 0.0709 & 0.1586 & \emph{0.2916} & 0.2504 & 0.5197 & \textbf{0.6356} \\
      & irrigation/aspiration HP & 0.9818 & 0.9848 & 0.9551 & 0.9652 & \emph{0.9854} & 0.9765 & 0.9846 & 0.9919 & 0.9954 & \textbf{0.9964} \\
      & phacoemulsifier HP       & 0.9877 & 0.9940 & 0.9813 & 0.9904 & 0.9923 & 0.9889 & \emph{0.9949} & 0.9967 & 0.9991 & \textbf{0.9992} \\
      & vitrectomy HP            & 0.4175 & 0.3869 & 0.6402 & 0.5496 & 0.4219 & \textbf{\emph{0.7296}} & 0.4154 & 0.6454 & 0.6088 & 0.6430 \\
      & implant injector         & 0.7701 & 0.7836 & 0.8331 & 0.8400 & \emph{0.8693} & 0.8026 & 0.8524 & 0.8965 & 0.9353 & \textbf{0.9386} \\
      & primary incision knife   & 0.7944 & 0.8644 & 0.8121 & 0.8443 & \emph{0.9081} & 0.7628 & 0.7823 & 0.9203 & 0.9696 & \textbf{0.9740} \\
      & secondary incision knife & 0.6524 & 0.8321 & 0.8434 & \emph{0.9195} & 0.9120 & 0.7946 & 0.8903 & 0.9169 & 0.9615 & \textbf{0.9649} \\
      & micromanipulator         & 0.9777 & 0.9843 & 0.9773 & 0.9786 & \emph{0.9878} & 0.9767 & 0.9867 & 0.9920 & 0.9950 & \textbf{0.9955} \\
      & suture needle            & 0.3740 & \emph{0.4728} & 0.3937 & 0.3957 & 0.4006 & 0.2942 & 0.3983 & 0.4702 & 0.8031 & \textbf{0.8204} \\
      & Mendez ring              & 0.1439 & \emph{0.8266} & 0.0759 & 0.0220 & 0.0083 & 0.0587 & 0.4292 & 0.3696 & 0.9606 & \textbf{0.9977} \\
      & Vannas scissors          & \emph{0.1987} & 0.1723 & 0.1284 & 0.0441 & 0.0930 & 0.0713 & 0.0760 & 0.2127 & 0.1937 & \textbf{0.2456} \\
      \cline{2-12}
      & Average (mAP)            & 0.5293 & 0.5899 & 0.5467 & 0.5529 & 0.5736 & 0.5532 & \emph{0.6086} & 0.6513 & 0.7734 & \textbf{0.7980} \\
      \cline{2-12}
      & Corr. with prevalence    & 0.6474 & 0.5560 & 0.5974 & 0.5918 & 0.5600 & 0.6382 & 0.6166 & 0.5544 & 0.4443 & 0.4329 \\
      \hline
      \multirow{8}{*}{\RotText{Cholec80}}
      & grasper                  & \emph{0.9723} & 0.9711 & 0.9621 & 0.9656 & 0.9627 & 0.9646 & 0.9711 & 0.9764 & \textbf{0.9767} & 0.9730 \\
      & bipolar                  & 0.9667 & 0.9643 & 0.9513 & 0.9491 & 0.9452 & 0.9477 & \emph{0.9688} & 0.9740 & \textbf{0.9823} & 0.9781 \\
      & hook                     & \emph{0.9986} & \emph{0.9986} & 0.9977 & 0.9980 & 0.9971 & 0.9979 & \emph{0.9986} & 0.9990 & \textbf{0.9992} & 0.9961 \\
      & scissors                 & 0.8802 & 0.8810 & 0.8101 & 0.8241 & 0.8120 & 0.8072 & \emph{0.8993} & 0.9155 & 0.9465 & \textbf{0.9501} \\
      & clipper                  & 0.9729 & 0.9722 & 0.9490 & 0.9567 & 0.9326 & 0.9446 & \emph{0.9814} & 0.9860 & \textbf{0.9958} & 0.9952 \\
      & irrigator                & 0.9568 & \emph{0.9572} & 0.9213 & 0.9330 & 0.9178 & 0.9319 & 0.9561 & 0.9692 & \textbf{0.9781} & 0.9657 \\
      & specimen bag             & 0.9555 & 0.9551 & 0.9310 & 0.9365 & 0.9364 & 0.9453 & \emph{0.9605} & 0.9691 & \textbf{0.9735} & 0.9729 \\
      \cline{2-12}
      & Average (mAP)            & 0.9576 & 0.9571 & 0.9318 & 0.9376 & 0.9291 & 0.9342 & \emph{0.9623} & 0.9699 & \textbf{0.9789} & 0.9759 \\
      \cline{2-12}
      & Corr. with prevalence    & 0.5477 & 0.5529 & 0.5899 & 0.5867 & 0.6371 & 0.5741 & 0.5261 & 0.4964 & 0.3838 & 0.3845 \\
    \end{tabular}}
  \end{center}
  \label{tab:AP}
\end{table*}

\begin{figure*}[!t]
  \begin{center}
    \changed{\begin{tabular}{c}
      \subfloat[test video from CATARACTS]{\includegraphics[width=0.75\textwidth]{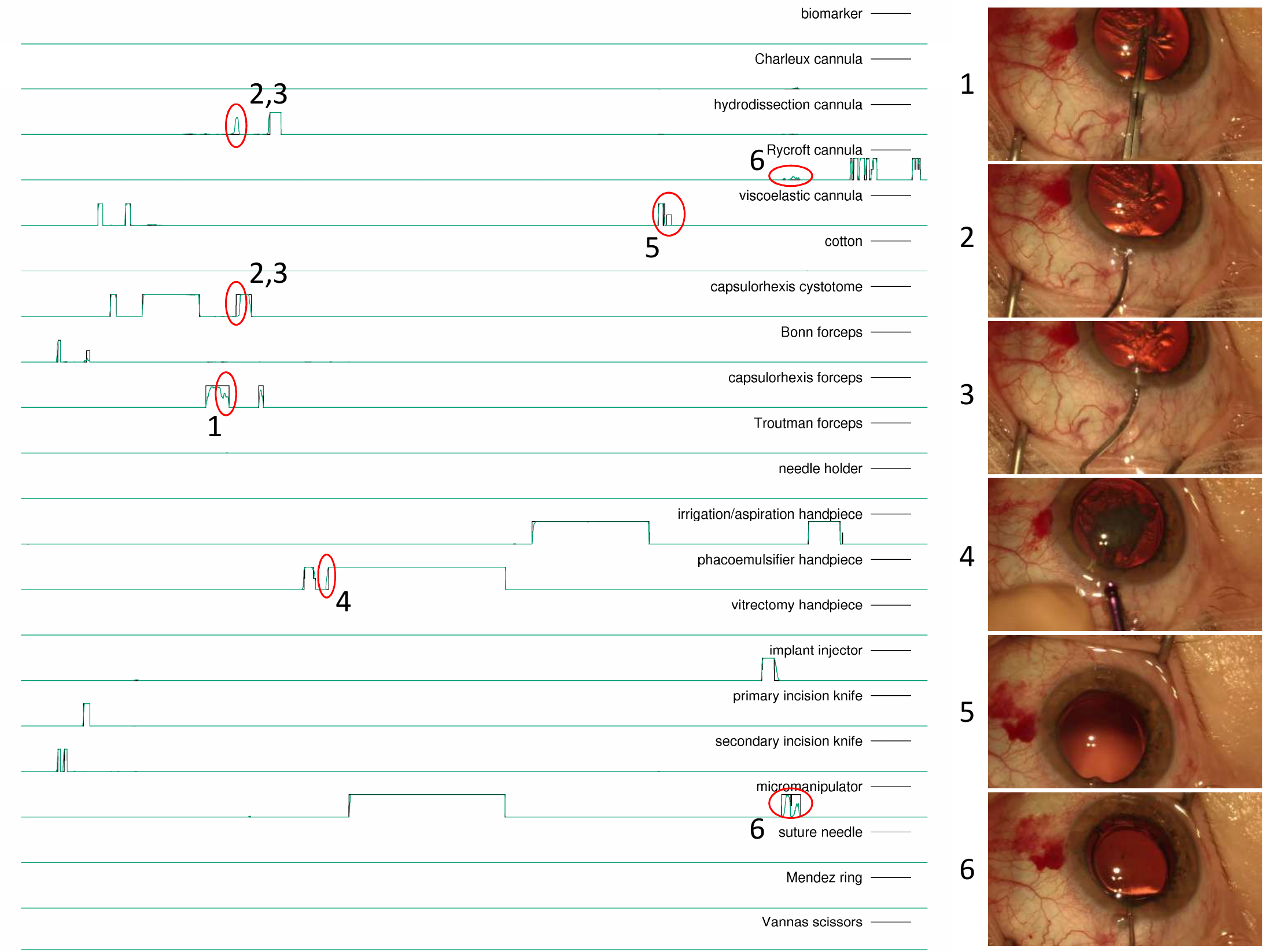}} \\
      \subfloat[test video from Cholec80]{\includegraphics[width=0.75\textwidth]{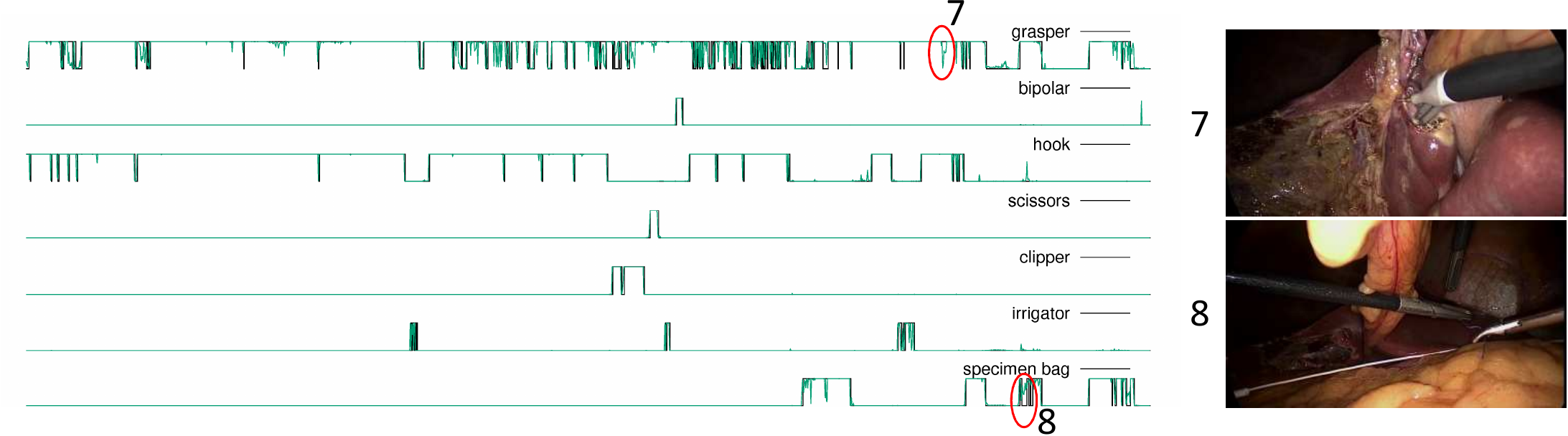}}
    \end{tabular}}
  \end{center}
  \caption{\changed{Sequence labeling for one test video from each dataset using the best ensemble of the ``all CNNs'' experiment (joint ``CNN+RNN'' Boosting, with median filtering for CATARACTS): tool usage according to human experts is in black, automatic predictions are in green. Areas surrounded by red circles are associated with images on the right.} \added{The label of image 1 (capsulorhexis forceps) has been correctly identified, but with a lower confidence level compared to previous images. The reason probably is that the forceps have remained closed for a long time and are therefore more difficult to recognize. In image 2, the capsulorhexis cystotome is detected as a hydrodissection cannula. The reason probably is that its distinctive claw-shaped tooltip is hidden in the incision and its distinctive elbow is out of the field of view. As soon as the elbow becomes visible (image 3), the correct label is assigned. In image 4, the phacoemulsifier handpiece is considered active, whereas it is not in contact with the eyeball yet. However, it touches the tear film, so the detector is almost correct. In image 5, one of the annotators indicated that the viscoelastic cannula is being used, although it is not actually visible: only indirect signs of presence (at the bottom) are visible; the detector was not able to recognize them. In image 6, the micromanipulator is partly mistaken for a Rycroft cannula: the explanation is similar for images 2 and 6. The reason why a hydrodissection cannula is detected in the former case and a Rycroft cannula in the latter probably comes from the RNN-based temporal modeling: hydrodissection cannulae are more likely at the beginning, Rycroft cannula are more likely at the end. In image 7, the grasper is not detected, probably because it is occluded by the hook. Finally, in image 8, a specimen bag is falsely detected, however the white string used for closing the bag is visible: the RNN-based temporal sequencer probably interpolated predictions from neighboring frames where the bag and the string are both visible.}}
  \label{fig:labeling_examples}
\end{figure*}

\begin{figure*}[!t]
  \begin{center}
    \includegraphics[width=0.75\textwidth]{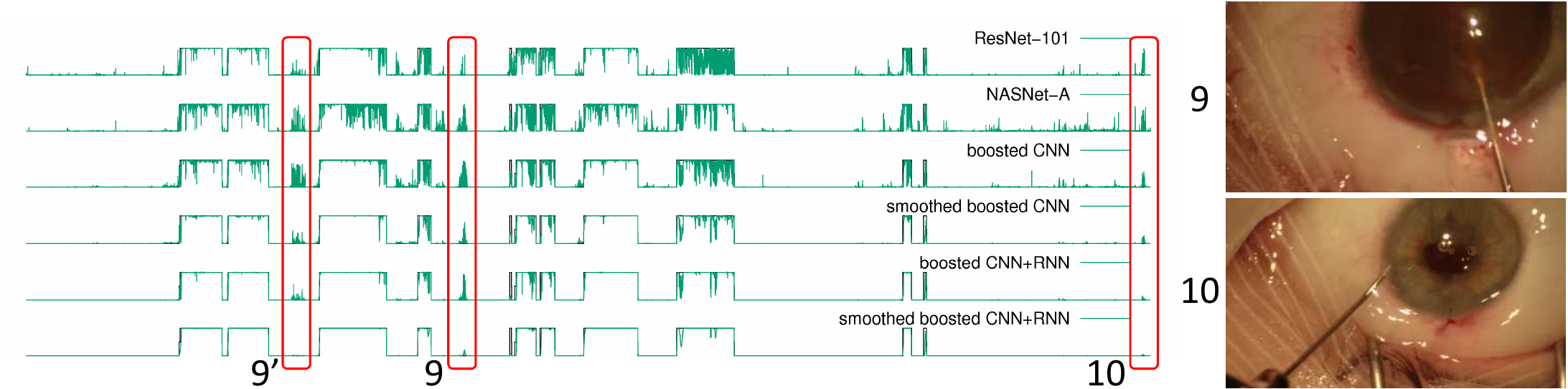}
  \end{center}
  \caption{\added{Sequence labeling for the micromanipulator tool for one test video from CATARACTS: tool usage according to human experts is in black, automatic predictions are in green. Areas surrounded by red circles are associated with images on the right. In images 9 and 9', the viscoelastic cannula is falsely detected as a micromanipulator. In image 10, a Rycroft cannula is falsely detected as a micromanipulator.}}
  \label{fig:labeling_comparison}
\end{figure*}

\changed{Sequence labeling examples obtained with the best ensembles are illustrated and commented in Fig. \ref{fig:labeling_examples}.} \added{In summary, mistakes made by the best ensembles are mainly due to occlusions. To illustrate the problems that the proposed ensemble solves, Fig. \ref{fig:labeling_comparison} reports labeling sequences obtained at different ensemble complexity levels. This figure suggests that the same errors are made by all detectors, but these errors are progressively attenuated as the ensemble becomes more complex.}

\subsection{\added{Comparisons with Baseline Solutions}}

\changed{The proposed ensemble (obtained through ``CNN+RNN boosting'', with median filtering for CATARACTS) is compared with various baseline methods in Table \ref{tab:comparisons}. For each baseline, the statistical significance of the difference with the proposed solution is assessed using a paired sample t-test.}
\begin{table*}[!t]
  \caption{\changed{Comparisons between the proposed ensemble (jointly boosted ``CNN+RNN'' ensemble, with median filtering for CATARACTS) and various baselines, in terms of mean area under the ROC curve (m$A_z$) and in terms of mean average precision (mAP). Non-significant differences at the 95\% confidence level are in bold. All CNNs are used to build ensembles unless specified otherwise (the weaker CNNs are Inception-ResNet-v2, ResNet-101 and ResNet-152). The last two experiments evaluate the proposed ensemble using the union or the intersection of tool usage annotations from both experts as ground truth.}}
  \begin{center}
    \footnotesize
    \changed{\begin{tabular}{l|cc|cc|cc}
                                                       & \multicolumn{2}{c|}{CATARACTS} & \multicolumn{2}{c|}{Cholec80}              & \multicolumn{2}{c}{p-value (paired sample t-test)} \\
      \cline{2-7}
                                                       & m$A_z$         & mAP            & m$A_z$         & mAP                      & difference in m$A_z$   & difference in mAP     \\
      \hline
      proposed ensemble                                & 0.9961         & 0.7980         & 0.9939         & 0.9789                   &                        &                       \\
      \hline
      boosted CNN                                      & 0.9916         & 0.6513         & 0.9923         & 0.9699                   & 2.964 $\times 10^{-4}$ & 0.001789              \\
      boosted CNN (weaker CNNs only)                   & 0.9829         & 0.6192         & 0.9880         & 0.9501                   & 2.479 $\times 10^{-4}$ & 0.003881              \\
      proposed ensemble (weaker CNNs only)             & 0.9900         & 0.6748         & 0.9917         & 0.9695                   & 0.007271               & 0.01401               \\
      smoothed ``sequentially'' boosted CNN+RNN        & 0.9939         & 0.6956         & 0.9930         & 0.9741                   & 0.002679               & 0.004047              \\
      proposed ensemble (unidirectional RNNs)          & 0.9957         & 0.7580         & 0.9936         & 0.9760                   & \textbf{0.05397}       & \textbf{0.07474}      \\
      \hline
      NASNet-A                                         & 0.9831         & 0.6086         & 0.9900         & 0.9623                   & 1.836 $\times 10^{-5}$ & 5.321 $\times 10^{-5}$\\
      NASNet-A + 1 LSTM                                & 0.9900         & 0.6949         & 0.9911         & 0.9723                   & 0.001088               & 0.001972              \\
      NASNet-A features + 1 LSTM                       & 0.9910         & 0.7264         & 0.9913         & 0.9755                   & 0.02051                & 0.01078               \\
      boosted CNN + smoothing                          & 0.9933         & 0.6735         & 0.9917         & 0.9703                   & 0.001317               & 0.003009              \\
      linear-combination CNN ensemble                  & 0.9913         & 0.6611         & 0.9917         & 0.9674                   & 5.328 $\times 10^{-4}$ & 0.001204              \\
      smoothed linear-combination CNN+LSTM ensemble    & 0.9937         & 0.7010         & 0.9926         & 0.9733                   & 0.002235               & 0.004602              \\
      \hline
      EndoNet \citep{twinanda_endonet:_2017}           & \multicolumn{2}{c|}{n/a}        & n/a            & 0.810                    & n/a                    & 0.005394              \\
      DResSys \citep{roychowdhury_identification_2017} & 0.9971         & n/a            & \multicolumn{2}{c|}{n/a}                  & \textbf{0.2936}        & n/a                   \\
      CUMV \citep{hu_surgical_2017}                    & 0.9897         & n/a            & \multicolumn{2}{c|}{n/a}                  & 0.001682               & n/a                   \\
      TROLIS \citep{marsalkaite_towards_2017}          & 0.9812         & n/a            & \multicolumn{2}{c|}{n/a}                  & 0.0204                 & n/a                   \\
      \hline
      proposed ensemble (union GT)                     & 0.9938         & 0.7876         & \multicolumn{2}{c|}{\multirow{2}{*}{n/a}} & 0.02266                & \textbf{0.2320}       \\
      proposed ensemble (intersection GT)              & 0.9958         & 0.7585         &                &                          & 0.001920               & 0.007045
    \end{tabular}}
  \end{center}
  \label{tab:comparisons}
\end{table*}

\changed{The first five baselines are variations on the proposed ensemble, as described above. All of these variations lead to decreased performance, with one exception: replacing bidirectional RNNs with unidirectional RNNs does not impact performance significantly. We note the good performance of ensembles obtained in the ``weaker CNNs'' experiment. In CATARACTS for instance, $A_z$ increases from 0.9663 for the best CNN (ResNet-152) to 0.9900 ($+ 0.0237$), while in the ``all CNNs'' experiment, it increases from 0.9831 to 0.9961 ($+ 0.0130$). On the downside, we also note that to achieve very high performance, good weak learners must be available.}

\changed{The next six baselines were proposed to evaluate the relevance of each part of the proposed framework. The first two tests show that only using the best CNN (NASNet-A in the ``all CNNs'' experiment) or the best CNN and the best RNN (NASNet-A + 1 LSTM) is clearly suboptimal. Interestingly, the third experiment shows that LSTMs operating on NASNet-A features (the 4032 features of the next to last NASNet-A layer) are better than LSTMs operating on NASNet-A predictions (the outputs of the last layer). However, besides being more computationally intensive, RNNs operating on CNN features are not compatible with boosting across multiple CNN architectures: feature layers would not necessarily have compatible shapes and could therefore not be combined linearly. The fourth experiment shows that the RNN part of the proposed ensemble cannot simply be replaced with a median filter. The ensemble evaluated in the fifth experiment is similar to the proposed boosted CNN ensemble, in the sense that predictions from several CNNs are combined linearly inside a sigmoid function. The difference is that each CNN (the seven CNNs studied in this paper) is trained independently; the weight assigned to each CNN is trained through a gradient descent. This approach is similar to the ensemble method proposed by \citet{roychowdhury_identification_2017}. The result of this experiment is rather disappointing: the performance of the resulting ensemble is almost as good as the boosted CNN ensemble ($p = 0.2390$ for $A_z$, $p = 0.7066$ for AP). The only advantage of the proposed ensemble is that it is more compact: four CNNs (see Fig. \ref{fig:boosting}) instead of seven. A similar ensemble is evaluated in the sixth experiment: one LSTM is trained independently on top of each of the seven CNNs and the predictions of these seven LSTMs is combined linearly inside a sigmoid function, again with weights obtained through a gradient descent. In CATARACTS, the ensemble predictions are then smoothed with a median filter. In that case, the proposed boosting approach is superior. We assume this superiority is mainly due to the proposed mechanism for boosting CNNs inside a ``CNN+RNN'' network (see section \ref{sec:BoostingCNNsInsideCNNRNNNetwork}), since the performance of the linear-combination ensemble is close to that of the ``sequentially'' boosted ``CNN+RNN''.} Ideally, we would also compare the proposed solution with the end-to-end training of a ``CNN+RNN'' network, but the complexity of that model prevents any experimentation.

\added{The next four baselines are recent solutions from the literature: EndoNet is from the original Cholec80 paper, the other three solutions are the top-ranking solutions of the CATARACTS challenge. We can see that the proposed solution is better than three of these solutions (EndoNet, CUMV and TROLIS) and not significantly worse than the other one (DResSys). One advantage of the proposed solution compared to DResSys is that it is more lightweight (less CNNs processing smaller images). The other advantage is that its unidirectional version, which is not significantly different from DResSys neither ($p = 0.07525$), allows online video sequencing, while DResSys jointly analyzes batches of $\sim$20,000 frames.}

\added{The last two experiments reported in Table \ref{tab:comparisons} evaluate the impact of the criterion chosen to define the ground truth in CATARACTS (exclusion of frames without a consensus). In those experiments, the ground truth is defined either as the union or the intersection of both expert interpretations, using all frames in the test videos. We can see that using those evaluation criteria decreases performance, in part because the most challenging frames (where experts disagree) are included, in part because the ground truth is of lower quality (more uncertain).}

\subsection{\added{Sensitivity} Analysis of the Boosted Video Labelers}

\begin{figure*}[!t]
  \begin{center}
    \includegraphics[width=0.96\textwidth]{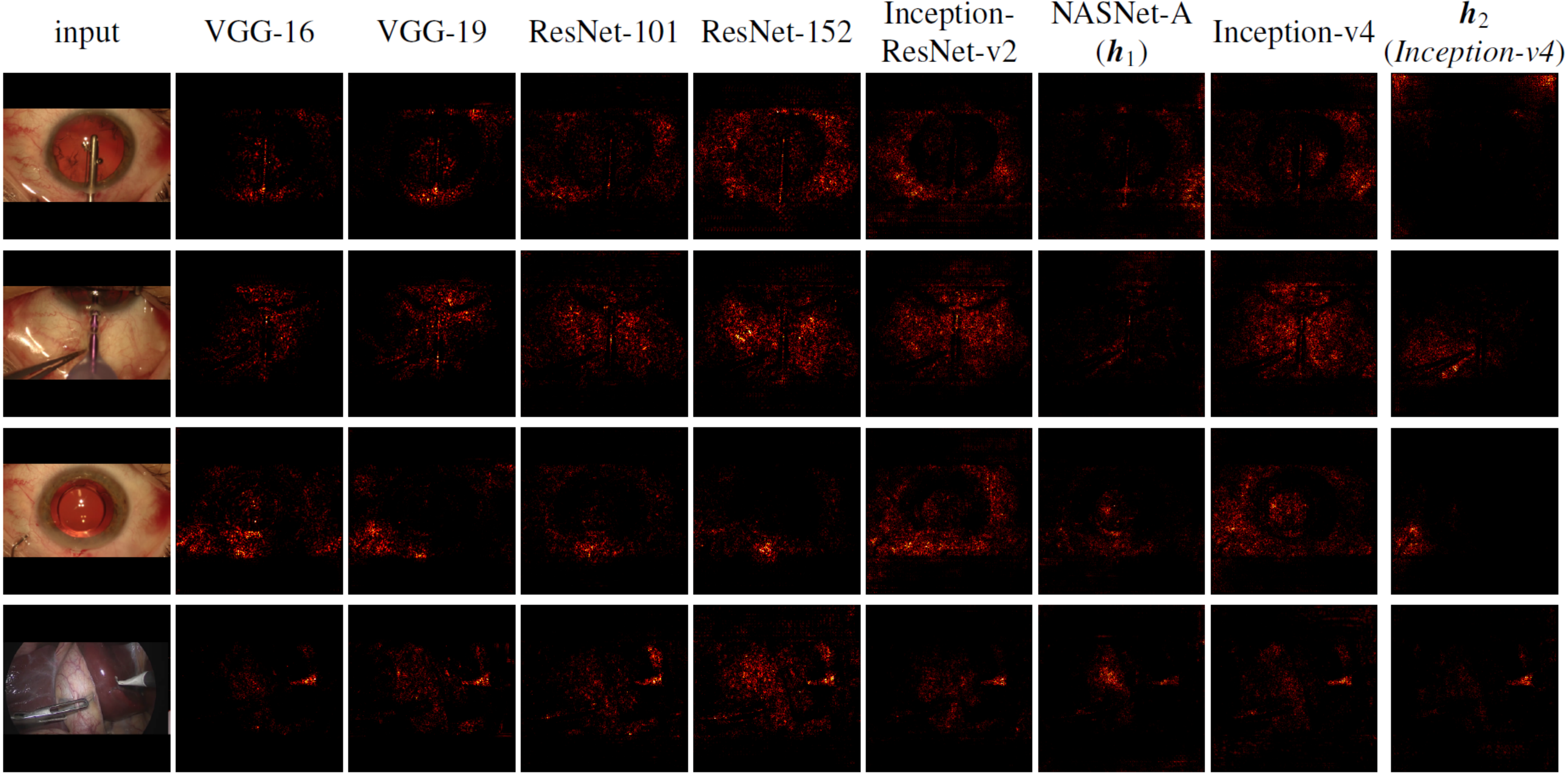}
  \end{center}
  \caption{\changed{Hue-constrained sensitivity analysis for multiple CNNs. The first three examples were taken from the the test set of CATARACTS. The last example was taken from the the test set of Cholec80. $\boldsymbol{h}_1$ and $\boldsymbol{h}_2$ are the first two CNNs selected in the ``all CNNs'' experiment.}}
  \label{fig:heatmaps}
\end{figure*}

To visualize what the CNNs have learned, one can rely on sensitivity analysis \citep{simonyan_deep_2014} and related metrics. Sensitivity is the gradient of the CNN predictions with respect to the pixel values: the pixel values influencing most the CNN predictions are highlighted. \changed{Recently, we proposed a variation on sensitivity called hue-constrained sensitivity} \citep{quellec_deep_2017}: the interpretation is similar, except that the three color components of a pixel are analyzed jointly rather than independently. \added{Given a CNN $\boldsymbol{h}$ and an input image $I$ with dimensions $W\times H\times 3$, the hue-constrained sensitivity heatmap $\pi$ of $I$ for $\boldsymbol{h}$ is defined as:
\begin{equation}
  \pi_{x,y} = \left| \frac{\partial \sum_{\theta\in\Theta}{h\left( m * I, \theta \right)}}{\partial m_{x,y}} \right|  \;\;,
  \label{eq:hueConstrainedSensitivityCriterion}
\end{equation}
where tensor $m$ is a matrix of ones with dimensions $W\times H$ and where '*' denotes the element-wise tensor multiplication. It should be noted that $m * I = I$ and that all color components of a pixel in $I$ are multiplied by the same tensor element in $m$, which ensures the desired hue preservation property \citep{quellec_deep_2017}.} Fig. \ref{fig:heatmaps} reports hue-constrained sensitivity heatmaps for \changed{all seven CNNs. It also reports heatmaps for $\boldsymbol{h}_2$, the second CNN (based on Inception-v4) added to the strong classifier in the ``all CNNs'' experiment (where $\boldsymbol{h}_1$ is NASNet-A).} This figure shows that, \added{in CATARACTS,} CNNs do not consider solely the tools, but also the anterior segment of the eye: the lens, which is modified by tools, the cornea, which is temporarily deformed by tools as they move, and the corneoscleral junction, where tools are inserted. One explanation is that each tool interacts differently with the eye and, therefore, analyzing the eye structures helps differentiating tools. Another explanation is that, \added{in this dataset,} the target labels are not related to tool presence, but rather to tool usage. So CNNs must be able to recognize whenever each tool is in contact with the eye. \added{This hypothesis is backed up by the observation that responses from tissues are lower in Cholec80, where tool usage is simply defined as tool visibility. We notice, however, that the best CNNs (NASNet-A and VGG-19) have sparser heatmaps and that those heatmaps are more focused on the tools. Heatmaps obtained for $\boldsymbol{h}_1$ (i.e NASNet-A) and $\boldsymbol{h}_2$ have been analyzed jointly to assess their complementarity. Because the first image was already classified well by NASNet-A, the heatmap for $\boldsymbol{h}_2$ is empty: the detections we see at the corner are just amplified noise (heatmap intensities have been normalized between 0 and 255). Similarly, in the second image, the phacoemulsifier handpiece at the center was detected well by NASNet-A, but not the forceps on the left: $\boldsymbol{h}_2$ seems to focus on the forceps. In the third image, we note that NASNet-A did not focus primarily on the tool (it seemed disturbed by specular reflections) but $\boldsymbol{h}_2$ does. In the last image, we also note a more focused heatmap for $\boldsymbol{h}_2$, compared to NASNet-A, although the grasper on the left (which was correctly detected by $\boldsymbol{h}_1$) is not detected anymore. So, overall, $\boldsymbol{h}_1$  and $\boldsymbol{h}_2$ are indeed complementary. And, clearly, the heatmaps for Inception-v4 before and after a boosting step are very different.}

\begin{figure*}[!t]
  \begin{center}
    \includegraphics[scale=.8]{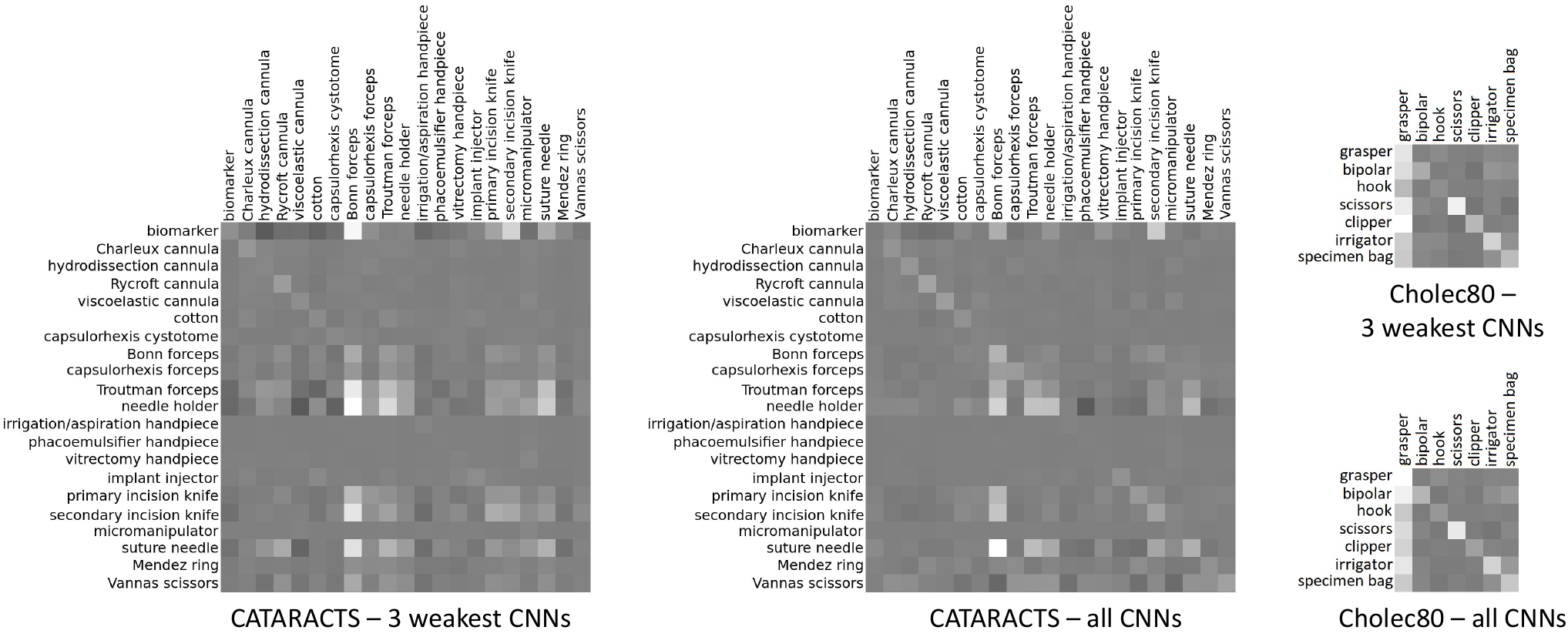}
  \end{center}
  \caption{\changed{Sensitivity analysis for $\boldsymbol{h}'_1$, the first added RNN in the two experiments based on joint ``CNN+RNN'' boosting: the ``all CNNs'' or ``3 weakest CNNs'' experiments. Intensity is proportional to $\nabla_{\phi,\theta}(\boldsymbol{h}'_1)$ [see Eq. (\ref{eq:nabla})]: gray means zero, black means negative, white means positive. Rows represent $\phi$, the label index in RNN predictions. Columns represent $\theta$, the label index in CNN predictions.}}
  \label{fig:gradients}
\end{figure*}
Because our joint ``CNN+RNN'' boosting algorithm relies on the gradients of RNN predictions with respect to CNN predictions [see Eq. (\ref{eq:valueOmegaPrime})], sensitivity analysis is also useful for RNNs in our case. These gradients are illustrated in a condensed form in Fig. \ref{fig:gradients}: given an RNN $\boldsymbol{h}'$, this figure shows $\nabla_{\phi,\theta}(\boldsymbol{h}')$, where:
\begin{equation}
  \nabla_{\phi,\theta}(\boldsymbol{h}') = \sum_{\boldsymbol{V}\in\mathcal{D}} \sum_{t} \sum_{u}{\frac{\partial h'(V_u, \phi)}{\partial p_L(V_t, \theta)}} \;.
  \label{eq:nabla}
\end{equation}
For a lazy RNN, all coefficients outside the diagonal would be zero. Here, we observe that the diagonal is not even \added{always} dominant. This is particularly true for tools \changed{whose detection performance increases greatly after RNN boosting}, such as needle holders, \changed{suture needles or Cholec80's scissors} (see Tables \ref{tab:Az} \added{and \ref{tab:AP}}): the gradients of RNN predictions for those tools with respect to CNN predictions for other tools are very high. Clearly, RNNs are not lazy and quite useful for this task.

\section{Discussion and Conclusions}
\label{sec:DiscussionConclusions}

A solution for labeling tool usage in cataract \added{and cholecystectomy} surgery videos has been presented. Following state-of-the-art video analysis solutions, it relies on convolutional neural networks (CNNs) for analyzing each frame in the video and on recurrent neural networks (RNNs) for analyzing the temporal sequencing throughout the entire surgery, based on the outputs of the CNNs. A novel framework for boosting a sequence labeler composed of CNNs and RNNs has been presented. The main motivation for this framework is the fact that ``CNN+RNN'' labelers cannot be trained from end to end, for complexity reasons. The framework allows to progressively improve the CNN and RNN parts of the system by adding weak classifiers (CNNs or RNNs) designed to improve the overall classification accuracy of the join system. In particular, like the theoretical end-to-end training solution, CNN training is supervised based on the outputs of the RNN block.

The proposed framework has several novelties. The main novelty lies in the boosting algorithm. CNN boosting had been proposed for multiclass classification problems \citep{moghimi_boosted_2016}. We adapted it for multilabel classification, showed its applicability to RNN boosting and, more importantly, introduced CNN boosting supervised based on the outputs of the RNN block. A second novelty lies in the proposed temporal sequence augmentation strategy: although very simple, it proved to be quite effective (see Fig. \ref{fig:subsamplingRate}).

The proposed framework is quite general and is likely applicable outside the scope of surgery video analysis. However, it is of particular relevance for this application because many tools are very similar to one another (e.g. the cannulae or the forceps --- see Fig. \ref{fig:tools}) but they are often used in a predefined order: using the temporal context (e.g. which tools have be used previously) is quite relevant for differentiating them. Therefore, it seems particularly useful to guide CNN training or boosting based on the temporal context. Experiments on \changed{two recent datasets (CATARACTS and Cholec80)} for the task of tool usage annotation demonstrated its very good performance: \changed{the mean area under the ROC curve reaches up to m$A_z = 0.9961$ over a collection of 21 cataract surgery tools and up to m$A_z = 0.9939$ over a collection of 7 cholecystectomy tools.}

\changed{If we look into the details of the proposed boosting solution, we first note that CNN boosting alone is disappointing: we found no significant difference between CNN boosting and a weighted sum of independently trained CNNs ($p = 0.2390$ for $A_z$, $p = 0.7066$ for AP), although the resulting architecture is more lightweight. The ability to boost CNNs based on the outputs of RNNs, on the other hand, leads to a significant improvement: joint ``CNN+RNN'' boosting is indeed significantly better than sequential ``CNN+RNN'' boosting ($p = 0.002679$ for $A_z$, $p = 0.004047$ for AP --- see Table \ref{tab:comparisons}).} Our explanation is that, when the CNN part is boosted independently of RNNs, much boosting effort is spent on trying to \changed{correct labeling errors, caused by previously selected CNNs, that RNNs could easily correct based on the temporal context}: using temporally-filtered outputs to supervise boosting makes more sense. These observations support our hypothesis that CNNs should be trained to be complimentary to RNNs.

\changed{One advantage of the proposed approach is that its online version, which relies on unidirectional RNNs, does not perform significantly worse than its offline version, relying on bidirectional RNNs ($p = 0.05397$ for $A_z$, $p = 0.07474$ for AP --- see Table \ref{tab:comparisons}). With slightly better performance, the offline version would be the preferred solution for report generation, surgical workflow optimization and surgical skill assessment. The online version, however, is the only valid solution for intraoperative warning or recommendation generation, provided that it is fast enough.} Similarly to the bidirectional version (see \changed{Fig. \ref{fig:boosting}}), the online version relies on \changed{three weak CNNs: one based on NASNet-A, one based on Inception-v4 and one based on VGG-16. All three together, processing one frame takes 50.9 ms using one GeForce GTX 1080 Ti GPU by Nvidia (see Table \ref{tab:computationTimes}).} Videos of the CATARACTS dataset have a frame rate of 30 image per second (i.e. 33.3 ms per image). It means a faster GPU would be required for real-time video analysis. Alternatively, two GPUs can be used, as the CNN classifiers can be run in parallel \added{(GPU 1: NASNet-A $\rightarrow$ 24.6 ms per image, GPU 2: Inception-v4 and VGG-16 $\rightarrow$ 26.3 ms per image). Note that the use of median filters (with radii of 32 frames at most) delays predictions by one second. In Cholec80, the frame rate is 1 image per second, so computation times are not an issue.}

\added{The proposed framework compares favorably with state-of-the-art competing solutions \citep{twinanda_endonet:_2017, hu_surgical_2017, marsalkaite_towards_2017}. In terms of $A_z$, it does not differ significantly from the winner of the CATARACTS challenge \citep{roychowdhury_identification_2017}. However, it has the advantage of being more lightweight and, more importantly, of allowing online video analysis.}

\added{This study has a few limitations. In particular, the same dataset was used to train CNNs and RNNs. Because CNN predictions are likely better in the learning set than in the validation and test sets, RNNs are trained under too favorable conditions, which could lead to overfitting. Because the number of learning videos is limited, we decided to use all of them for training CNNs and RNNs. We simply relied on early stopping to discard overfitted configurations. Another limitation is that we did not explore data rebalancing techniques \citep{sahu_addressing_2017} or weighted cost functions to deal with multi-label imbalance, assuming that boosting can deal with it satisfactorily.}

\added{In conclusion, an accurate solution for labeling tool usage in surgery videos has been presented.} In view of the good performance, automatic surgery monitoring can now be envisaged seriously \citep{charriere_real-time_2017}. \changed{We are currently exploring solutions to provide useful feedbacks to the surgeon,} based on information collected during the surgery. Support to beginners is a particular relevant application, but many more can be envisioned for the near future.


\bibliographystyle{elsarticle-harv}
\bibliography{cataracts}

\begin{thebibliography}{60}
\expandafter\ifx\csname natexlab\endcsname\relax\def\natexlab#1{#1}\fi
\expandafter\ifx\csname url\endcsname\relax
  \def\url#1{\texttt{#1}}\fi
\expandafter\ifx\csname urlprefix\endcsname\relax\def\urlprefix{URL }\fi

\bibitem[{Al~Hajj et~al.(2017)Al~Hajj, Lamard, Charri{\`e}re, Cochener, and
  Quellec}]{al_hajj_surgical_2017}
Al~Hajj, H., Lamard, M., Charri{\`e}re, K., Cochener, B., Quellec, G., 2017.
  Surgical tool detection in cataract surgery videos through multi-image fusion
  inside a convolutional neural network. In: Proc {IEEE} {EMBC}. Jeju Island,
  Korea.

\bibitem[{Bodenstedt et~al.(2017)Bodenstedt, Wagner, Kati{\'c}, Mietkowski,
  Mayer, Kenngott, M{\"u}ller-Stich, Dillmann, and
  Speidel}]{bodenstedt_unsupervised_2017}
Bodenstedt, S., Wagner, M., Kati{\'c}, D., Mietkowski, P., Mayer, B., Kenngott,
  H., M{\"u}ller-Stich, B., Dillmann, R., Speidel, S., Feb. 2017. Unsupervised
  temporal context learning using convolutional neural networks for
  laparoscopic workflow analysis. Tech. Rep. arXiv:1702.03684 [cs], Karlsruhe
  Institute of Technology.

\bibitem[{Bouget et~al.(2017)Bouget, Allan, Stoyanov, and
  Jannin}]{bouget_vision-based_2017}
Bouget, D., Allan, M., Stoyanov, D., Jannin, P., 2017. Vision-based and
  marker-less surgical tool detection and tracking: a review of the literature.
  Med Image Anal 35, 633--654.

\bibitem[{Cad{\`e}ne et~al.(2016)Cad{\`e}ne, Robert, Thome, and
  Cord}]{cadene_m2cai_2016}
Cad{\`e}ne, R., Robert, T., Thome, N., Cord, M., Oct. 2016. M2cai workflow
  challenge: convolutional neural networks with time smoothing and hidden
  {Markov} model for video frames classification. Tech. Rep. arXiv:1610.05541
  [cs], Universit{\'e} de Pierre et Marie Curie.

\bibitem[{Charri{\`e}re et~al.(2017)Charri{\`e}re, Quellec, Lamard, Martiano,
  Cazuguel, Coatrieux, and Cochener}]{charriere_real-time_2017}
Charri{\`e}re, K., Quellec, G., Lamard, M., Martiano, D., Cazuguel, G.,
  Coatrieux, G., Cochener, B., Nov. 2017. Real-time analysis of cataract
  surgery videos using statistical models. Multimed Tools Appl 76~(21),
  22473--22491.

\bibitem[{Chen et~al.(2017)Chen, Chen, Hu, Chen, and Wang}]{chen_action_2017}
Chen, H., Chen, J., Hu, R., Chen, C., Wang, Z., Mar. 2017. Action recognition
  with temporal scale-invariant deep learning framework. China Communications
  14~(2), 163--172.

\bibitem[{Cho et~al.(2014)Cho, van Merrienboer, Bahdanau, and
  Bengio}]{cho_properties_2014}
Cho, K., van Merrienboer, B., Bahdanau, D., Bengio, Y., Oct. 2014. On the
  properties of neural machine translation: {Encoder-decoder} approaches. In:
  Proc {SSST}. Doha, Qatar, pp. 103--111, arXiv: 1409.1259.

\bibitem[{Dergachyova et~al.(2016)Dergachyova, Bouget, Huaulm{\'e}, Morandi,
  and Jannin}]{dergachyova_automatic_2016}
Dergachyova, O., Bouget, D., Huaulm{\'e}, A., Morandi, X., Jannin, P., Jun.
  2016. Automatic data-driven real-time segmentation and recognition of
  surgical workflow. Int J Comput Assist Radiol Surg 11~(6), 1081--1089.

\bibitem[{Dipietro et~al.(2016)Dipietro, Lea, Malpani, Ahmidi, Vedula, Lee,
  Lee, and Hager}]{dipietro_recognizing_2016}
Dipietro, R., Lea, C., Malpani, A., Ahmidi, N., Vedula, S., Lee, G., Lee, M.,
  Hager, G., Oct. 2016. Recognizing surgical activities with recurrent neural
  networks. In: Proc {MICCAI}. Athens, Greece, pp. 551--558.

\bibitem[{Donahue et~al.(2017)Donahue, Hendricks, Rohrbach, Venugopalan,
  Guadarrama, Saenko, and Darrell}]{donahue_long-term_2017}
Donahue, J., Hendricks, L., Rohrbach, M., Venugopalan, S., Guadarrama, S.,
  Saenko, K., Darrell, T., Apr. 2017. Long-term recurrent convolutional
  networks for visual recognition and description. IEEE Trans Pattern Anal Mach
  Intell 39~(4), 677--691.

\bibitem[{Feng et~al.(2016)Feng, Li, and Luo}]{feng_learning_2016}
Feng, Y., Li, Y., Luo, J., Dec. 2016. Learning effective gait features using
  {LSTM}. In: Proc {IEEE} {ICPR}. Cancun, Mexico, pp. 325--330.

\bibitem[{Freund and Schapire(1997)}]{freund_decision-theoretic_1997}
Freund, Y., Schapire, R.~E., Aug. 1997. A {Decision}-theoretic generalization
  of on-line learning and an application to boosting. J Comput Syst Sci 55~(1),
  119--139.

\bibitem[{Friedman(2001)}]{friedman_greedy_2001}
Friedman, J.~H., Oct. 2001. Greedy function approximation: {A} gradient
  boosting machine. Ann Stat 29~(5), 1189--1232.

\bibitem[{Gammulle et~al.(2017)Gammulle, Denman, Sridharan, and
  Fookes}]{gammulle_two_2017}
Gammulle, H., Denman, S., Sridharan, S., Fookes, C., Mar. 2017. Two stream
  {LSTM}: {A} deep fusion framework for human action recognition. In: Proc
  {IEEE} {WACV}. Santa Rosa, CA, USA, pp. 177--186.

\bibitem[{Gao et~al.(2016)Gao, Rong, Shen, and Xiong}]{gao_convolutional_2016}
Gao, Y., Rong, W., Shen, Y., Xiong, Z., Jul. 2016. Convolutional neural network
  based sentiment analysis using {Adaboost} combination. In: Proc {IEEE}
  {IJCNN}. Vancouver, Canada, pp. 1333--1338.

\bibitem[{He et~al.(2016{\natexlab{a}})He, Zhang, Ren, and Sun}]{he_deep_2016}
He, K., Zhang, X., Ren, S., Sun, J., Jun. 2016{\natexlab{a}}. Deep residual
  learning for image recognition. In: Proc {CVPR}. Las Vegas, NV, USA, pp.
  770--778.

\bibitem[{He et~al.(2016{\natexlab{b}})He, Zhang, Ren, and
  Sun}]{he_identity_2016}
He, K., Zhang, X., Ren, S., Sun, J., Oct. 2016{\natexlab{b}}. Identity mappings
  in deep residual networks. In: Proc {ECCV}. Lecture {Notes} in {Computer}
  {Science}. Springer, Cham, Amsterdam, The Netherlands, pp. 630--645.

\bibitem[{Hochreiter and Schmidhuber(1997)}]{hochreiter_long_1997}
Hochreiter, S., Schmidhuber, J., Nov. 1997. Long short-term memory. Neural
  Comput 9~(8), 1735--1780.

\bibitem[{Hu and Heng(2017)}]{hu_surgical_2017}
Hu, X., Heng, P.-A., Nov. 2017. Surgical tool annotation in cataract surgery
  videos. Tech. rep., Chinese University of Hong Kong.

\bibitem[{Huang et~al.(2017)Huang, Liu, Maaten, and
  Weinberger}]{huang_densely_2017}
Huang, G., Liu, Z., Maaten, L. v.~d., Weinberger, K.~Q., Jul. 2017. Densely
  connected convolutional networks. In: Proc {IEEE} {CVPR}. Honolulu, HI, USA,
  pp. 2261--2269.

\bibitem[{Ji et~al.(2013)Ji, Xu, Yang, and Yu}]{ji_3d_2013}
Ji, S., Xu, W., Yang, M., Yu, K., Jan. 2013. {3D} convolutional neural networks
  for human action recognition. IEEE Trans Pattern Mach Intell 35~(1),
  221--231.

\bibitem[{Jin et~al.(2016)Jin, Dou, Chen, Yu, and Heng}]{jin_endorcn:_2016}
Jin, Y., Dou, Q., Chen, H., Yu, L., Heng, P.-A., Oct. 2016. {EndoRCN}:
  recurrent convolutional networks for recognition of surgical workflow in
  cholecystectomy procedure video. Tech. rep., The Chinese University of Hong
  Kong.

\bibitem[{Khorrami et~al.(2016)Khorrami, Le, Brady, Dagli, and
  Huang}]{khorrami_how_2016}
Khorrami, P., Le, P., Brady, K., Dagli, C., Huang, T., Sep. 2016. How deep
  neural networks can improve emotion recognition on video data. In: Proc
  {IEEE} {ICIP}. Phoenix, AZ, USA, pp. 619--623.

\bibitem[{Krizhevsky et~al.(2012)Krizhevsky, Sutskever, and
  Hinton}]{krizhevsky_imagenet_2012}
Krizhevsky, A., Sutskever, I., Hinton, G.~E., Dec. 2012. {ImageNet}
  classification with deep convolutional neural networks. In: Proc NIPS.
  Vol.~25. Granada, Spain, pp. 1097--1105.

\bibitem[{Lalys and Jannin(2014)}]{lalys_surgical_2014}
Lalys, F., Jannin, P., May 2014. Surgical process modelling: a review. Int J
  Comput Assist Radiol Surg 9~(3), 495--511.

\bibitem[{Lea et~al.(2016{\natexlab{a}})Lea, Vidal, and
  Hager}]{lea_learning_2016}
Lea, C., Vidal, R., Hager, G.~D., May 2016{\natexlab{a}}. Learning
  convolutional action primitives for fine-grained action recognition. In: Proc
  {IEEE} {ICRA}. Stockholm, Sweden, pp. 1642--1649.

\bibitem[{Lea et~al.(2016{\natexlab{b}})Lea, Vidal, Reiter, and
  Hager}]{lea_temporal_2016}
Lea, C., Vidal, R., Reiter, A., Hager, G., Oct. 2016{\natexlab{b}}. Temporal
  convolutional networks: a unified approach to action segmentation. In: Proc
  {ECCV}. Amsterdam, The Netherlands, pp. 47--54.

\bibitem[{Mar\v{s}alkait\.{e} et~al.(2017)Mar\v{s}alkait\.{e},
  Bialopetravi\v{c}ius, and Armaitis}]{marsalkaite_towards_2017}
Mar\v{s}alkait\.{e}, G., Bialopetravi\v{c}ius, J., Armaitis, J., Nov. 2017.
  Towards robust tool identification for cataract surgery. Tech. rep., Oxipit,
  UAB.

\bibitem[{Mason et~al.(1999)Mason, Baxter, Bartlett, and
  Frean}]{mason_boosting_1999}
Mason, L., Baxter, J., Bartlett, P.~L., Frean, M.~R., Dec. 1999. Boosting
  algorithms as gradient descent. In: Proc {NIPS}. Vol.~12. Denver, CO, USA,
  pp. 512--518.

\bibitem[{Mishra et~al.(2017)Mishra, Sathish, and Sheet}]{mishra_learning_2017}
Mishra, K., Sathish, R., Sheet, D., Jul. 2017. Learning latent temporal
  connectionism of deep residual visual abstractions for identifying surgical
  tools in laparoscopy procedures. In: Proc {IEEE} {CVPR} {Works}. Honolulu,
  HI, USA, pp. 2233--2240.

\bibitem[{Moghimi et~al.(2016)Moghimi, Saberian, Yang, Li, Vasconcelos, and
  Belongie}]{moghimi_boosted_2016}
Moghimi, M., Saberian, M., Yang, J., Li, L.-J., Vasconcelos, N., Belongie, S.,
  Sep. 2016. Boosted convolutional neural networks. In: Proc {BMVC}. York, UK.

\bibitem[{Primus et~al.(2018)Primus, Putzgruber-Adamitsch, Taschwer,
  M{\"u}nzer, El-Shabrawi, B{\"o}sz{\"o}rmenyi, and
  Schoeffmann}]{primus_frame-based_2018}
Primus, M., Putzgruber-Adamitsch, D., Taschwer, M., M{\"u}nzer, B.,
  El-Shabrawi, Y., B{\"o}sz{\"o}rmenyi, L., Schoeffmann, K., Feb. 2018.
  Frame-based classification of operation phases in cataract surgery videos.
  In: Proc {MMM}. Vol. 10704 LNCS. Bangkok, Thailand, pp. 241--253.

\bibitem[{Quellec et~al.(2017)Quellec, Charri{\`e}re, Boudi, Cochener, and
  Lamard}]{quellec_deep_2017}
Quellec, G., Charri{\`e}re, K., Boudi, Y., Cochener, B., Lamard, M., Jul. 2017.
  Deep image mining for diabetic retinopathy screening. Med Image Anal 39,
  178--193.

\bibitem[{Quellec et~al.(2014)Quellec, Lamard, Cochener, and
  Cazuguel}]{quellec_real-time_2014}
Quellec, G., Lamard, M., Cochener, B., Cazuguel, G., Dec. 2014. Real-time
  segmentation and recognition of surgical tasks in cataract surgery videos.
  IEEE Trans Med Imaging 33~(12), 2352--2360.

\bibitem[{Quellec et~al.(2015)Quellec, Lamard, Cochener, and
  Cazuguel}]{quellec_real-time_2015}
Quellec, G., Lamard, M., Cochener, B., Cazuguel, G., Apr. 2015. Real-time task
  recognition in cataract surgery videos using adaptive spatiotemporal
  polynomials. IEEE Trans Med Imaging 34~(4), 877--887.

\bibitem[{Raju et~al.(2016)Raju, Wang, and Huang}]{raju_m2cai_2016}
Raju, A., Wang, S., Huang, J., Oct. 2016. {M2CAI} surgical tool detection
  challenge report. Tech. rep., University of Texas at Arlington.

\bibitem[{Roychowdhury et~al.(2017)Roychowdhury, Bian, Vahdat, and
  William~G.}]{roychowdhury_identification_2017}
Roychowdhury, S., Bian, Z., Vahdat, A., William~G., M., Nov. 2017.
  Identification of surgical tools using deep neural networks. Tech. rep.,
  D-Wave Systems Inc.

\bibitem[{Russakovsky et~al.(2015)Russakovsky, Deng, Su, Krause, Satheesh, Ma,
  Huang, Karpathy, Khosla, Bernstein, Berg, and
  Fei-Fei}]{russakovsky_imagenet_2015}
Russakovsky, O., Deng, J., Su, H., Krause, J., Satheesh, S., Ma, S., Huang, Z.,
  Karpathy, A., Khosla, A., Bernstein, M., Berg, A.~C., Fei-Fei, L., Apr. 2015.
  {ImageNet} large scale visual recognition challenge. Int J Comput Vis
  115~(3), 211--252.

\bibitem[{Sahu et~al.(2016)Sahu, Mukhopadhyay, Szengel, and
  Zachow}]{sahu_tool_2016}
Sahu, M., Mukhopadhyay, A., Szengel, A., Zachow, S., Oct. 2016. Tool and phase
  recognition using contextual {CNN} features. Tech. Rep. arXiv:1610.08854
  [cs.CV], Zuse Institute Berlin.

\bibitem[{Sahu et~al.(2017)Sahu, Mukhopadhyay, Szengel, and
  Zachow}]{sahu_addressing_2017}
Sahu, M., Mukhopadhyay, A., Szengel, A., Zachow, S., Jun. 2017. Addressing
  multi-label imbalance problem of surgical tool detection using {CNN}. Int J
  Comput Assist Radiol Surg 12~(6), 1013--1020.

\bibitem[{Schuster and Paliwal(1997)}]{schuster_bidirectional_1997}
Schuster, M., Paliwal, K.~K., Nov. 1997. Bidirectional recurrent neural
  networks. IEEE Trans Signal Process 45~(11), 2673--2681.

\bibitem[{Shen et~al.(2017)Shen, Wu, and Suk}]{shen_deep_2017}
Shen, D., Wu, G., Suk, H.-I., Jun. 2017. Deep learning in medical image
  analysis. Annu Rev Biomed Eng 19, 221--248.

\bibitem[{Simonyan et~al.(2014)Simonyan, Vedaldi, and
  Zisserman}]{simonyan_deep_2014}
Simonyan, K., Vedaldi, A., Zisserman, A., Apr. 2014. Deep inside convolutional
  networks: visualising image classification models and saliency maps. In:
  {ICLR} {Workshop}. Calgary, Canada.

\bibitem[{Simonyan and Zisserman(2014)}]{simonyan_two-stream_2014}
Simonyan, K., Zisserman, A., Dec. 2014. Two-stream convolutional networks for
  action recognition in videos. In: Proc {NIPS}. Vol.~27. Montreal, Canada, pp.
  568--576.

\bibitem[{Simonyan and Zisserman(2015)}]{simonyan_very_2015}
Simonyan, K., Zisserman, A., May 2015. Very deep convolutional networks for
  large-scale image recognition. In: Proc {ICLR}. San Diego, CA, USA.

\bibitem[{Singh et~al.(2016)Singh, Marks, Jones, Tuzel, and
  Shao}]{singh_multi-stream_2016}
Singh, B., Marks, T., Jones, M., Tuzel, O., Shao, M., Jan. 2016. A multi-stream
  bi-directional recurrent neural network for fine-grained action detection.
  In: Proc {IEEE} {CVPR}. Las Vegas, NV, USA, pp. 1961--1970.

\bibitem[{Szegedy et~al.(2017)Szegedy, Ioffe, Vanhoucke, and
  Alemi}]{szegedy_inception-v4_2017}
Szegedy, C., Ioffe, S., Vanhoucke, V., Alemi, A., Feb. 2017. Inception-v4,
  {Inception}-{ResNet} and the impact of residual connections on learning. In:
  Proc {AAAI}. San Francisco, CA, USA, pp. 4278--4284.

\bibitem[{Szegedy et~al.(2015{\natexlab{a}})Szegedy, Liu, Jia, Sermanet, Reed,
  Anguelov, Erhan, Vanhoucke, and Rabinovich}]{szegedy_going_2015}
Szegedy, C., Liu, W., Jia, Y., Sermanet, P., Reed, S., Anguelov, D., Erhan, D.,
  Vanhoucke, V., Rabinovich, A., Jun. 2015{\natexlab{a}}. Going deeper with
  convolutions. In: Proc {IEEE} {CVPR}. Boston, MA, USA, pp. 1--9.

\bibitem[{Szegedy et~al.(2015{\natexlab{b}})Szegedy, Vanhoucke, Ioffe, Shlens,
  and Wojna}]{szegedy_rethinking_2015}
Szegedy, C., Vanhoucke, V., Ioffe, S., Shlens, J., Wojna, Z., Dec.
  2015{\natexlab{b}}. Rethinking the {Inception} architecture for computer
  vision. Tech. Rep. arXiv:1512.00567 [cs], Google.

\bibitem[{Tao et~al.(2013)Tao, Zappella, Hager, and Vidal}]{tao_surgical_2013}
Tao, L., Zappella, L., Hager, G.~D., Vidal, R., Sep. 2013. Surgical gesture
  segmentation and recognition. In: Proc {MICCAI}. Nagoya, Japan, pp. 339--346.

\bibitem[{Tran et~al.(2017)Tran, Sakurai, Yamazoe, and Lee}]{tran_phase_2017}
Tran, D., Sakurai, R., Yamazoe, H., Lee, J.-H., 2017. Phase segmentation
  methods for an automatic surgical workflow analysis. Int J Biomed Imaging
  2017, 1985796.

\bibitem[{Trikha et~al.(2013)Trikha, Turnbull, Morris, Anderson, and
  Hossain}]{trikha_journey_2013}
Trikha, S., Turnbull, A. M.~J., Morris, R.~J., Anderson, D.~F., Hossain, P.,
  Apr. 2013. The journey to femtosecond laser-assisted cataract surgery: new
  beginnings or a false dawn? Eye (Lond) 27~(4), 461--473.

\bibitem[{Twinanda et~al.(2016)Twinanda, Mutter, Marescaux, de~Mathelin, and
  Padoy}]{twinanda_single-_2016}
Twinanda, A.~P., Mutter, D., Marescaux, J., de~Mathelin, M., Padoy, N., Oct.
  2016. Single- and multi-task architectures for surgical workflow challenge at
  {M}2cai 2016. Tech. Rep. arXiv:1610.08844 [cs], University of Strasbourg.

\bibitem[{Twinanda et~al.(2017)Twinanda, Shehata, Mutter, Marescaux,
  de~Mathelin, and Padoy}]{twinanda_endonet:_2017}
Twinanda, A.~P., Shehata, S., Mutter, D., Marescaux, J., de~Mathelin, M.,
  Padoy, N., Jan. 2017. {EndoNet}: {A} deep architecture for recognition tasks
  on laparoscopic videos. IEEE Trans Med Imaging 36~(1), 86--97.

\bibitem[{Walach and Wolf(2016)}]{walach_learning_2016}
Walach, E., Wolf, L., Oct. 2016. Learning to count with {CNN} boosting. In:
  Proc {ECCV}. Vol. 9906. Amsterdam, The Netherlands, pp. 660--676.

\bibitem[{Wang et~al.(2017)Wang, Gao, Song, and Shen}]{wang_beyond_2017}
Wang, X., Gao, L., Song, J., Shen, H., Apr. 2017. Beyond frame-level {CNN}:
  saliency-aware {3D} {CNN} with {LSTM} for video action recognition. IEEE
  Signal Processing Letters 24~(4), 510--514.

\bibitem[{Zappella et~al.(2013)Zappella, B{\'e}jar, Hager, and
  Vidal}]{zappella_surgical_2013}
Zappella, L., B{\'e}jar, B., Hager, G., Vidal, R., Oct. 2013. Surgical gesture
  classification from video and kinematic data. Med Image Anal 17~(7),
  732--745.

\bibitem[{Zhang et~al.(2016)Zhang, Du, and Zhang}]{zhang_scene_2016}
Zhang, F., Du, B., Zhang, L., Mar. 2016. Scene classification via a gradient
  boosting random convolutional network framework. IEEE Trans Geosci Remote
  Sens 54~(3), 1793--1802.

\bibitem[{Zia et~al.(2016)Zia, Castro, and Essa}]{zia_fine-tuning_2016}
Zia, A., Castro, D., Essa, I., Oct. 2016. Fine-tuning deep architectures for
  surgical tool detection. Tech. rep., Georgia Institute of Technology.

\bibitem[{Zoph et~al.(2017)Zoph, Vasudevan, Shlens, and
  Le}]{zoph_learning_2017}
Zoph, B., Vasudevan, V., Shlens, J., Le, Q.~V., Jul. 2017. Learning
  transferable architectures for scalable image recognition. arXiv:1707.07012
  [cs, stat].

\end{thebibliography}

\end{document}